\documentclass{article}
\pdfoutput=1

\usepackage{iclr2027_conference,times}
\usepackage{amsmath,amsfonts,bm}

\def\eqref#1{equation~\ref{#1}}

\def\1{\bm{1}}

\DeclareMathAlphabet{\mathsfit}{\encodingdefault}{\sfdefault}{m}{sl}
\SetMathAlphabet{\mathsfit}{bold}{\encodingdefault}{\sfdefault}{bx}{n}

\usepackage[T1]{fontenc}
\usepackage[utf8]{inputenc}

\usepackage[table]{xcolor}
\usepackage{colortbl}
\usepackage[normalem]{ulem}

\definecolor{text_high}{RGB}{255,180,180}
\definecolor{text_second_high}{RGB}{255,230,230}
\definecolor{figure_high}{RGB}{180,255,180}
\definecolor{figure_second_high}{RGB}{230,255,230}
\definecolor{customred}{RGB}{255,0,0}

\usepackage{arydshln}
\usepackage{tabularx}
\usepackage{booktabs}
\usepackage{multirow}
\usepackage{graphicx}
\usepackage{float}
\usepackage{placeins}

\usepackage{amsmath}
\usepackage{amssymb}
\usepackage{bm}

\usepackage{microtype}
\usepackage{xspace}
\usepackage{inconsolata}
\usepackage{fontawesome5}

\usepackage{pifont}
\usepackage{enumitem}

\usepackage[most]{tcolorbox}
\newtcolorbox{blueBox}[1][]{
  colback=blue!5!white,
  colframe=blue!75!black,
  title=#1,
  fonttitle=\bfseries,
  boxrule=0.7pt,
  arc=4pt,
  left=3pt, right=3pt, top=3pt, bottom=3pt
}
\newtcolorbox{wronganswer}[1][]{
  enhanced, breakable,
  colframe=customred,
  colback=customred!10!white,
  sharp corners,
  boxsep=0pt,
  left=5pt, right=5pt, top=6pt, bottom=6pt,
  boxrule=0pt, leftrule=4pt,
  #1
}

\usepackage{eulervm}
\usepackage{wrapfig}
\usepackage{caption}

\usepackage{fancyvrb}

\definecolor{promptframe}{HTML}{6B3A2E}
\definecolor{inputback}{HTML}{3A3A3A}
\definecolor{inputfield}{HTML}{ECECEC}

\newtcolorbox{promptbox}[1]{%
  enhanced, colback=white, colframe=promptframe,
  boxrule=0.9pt, arc=2pt, left=6pt, right=6pt, top=6pt, bottom=6pt,
  fonttitle=\footnotesize\bfseries\sffamily, coltitle=white, title={#1},
  attach boxed title to top left={xshift=0mm, yshift=-0.1mm},
  boxed title style={colback=promptframe, colframe=promptframe,
                     arc=1pt, boxrule=0pt, left=4pt, right=4pt}}

\newtcolorbox{imginput}{%
  enhanced, colback=inputfield, colframe=inputback, boxrule=0.8pt, arc=1pt,
  left=5pt, right=5pt, top=3pt, bottom=3pt,
  fonttitle=\scriptsize\bfseries\sffamily, coltitle=white, title={Image input},
  attach boxed title to top left={xshift=0mm, yshift=-0.1mm},
  boxed title style={colback=inputback, colframe=inputback, boxrule=0pt,
                     arc=1pt, left=4pt, right=4pt}}

\newcommand{\ours}{\textsc{DR Pixel}\xspace}
\newcommand{\oursbase}{\textsc{DR Pixel}$_\text{Base}$\xspace}
\newcommand{\ourstrained}{\textsc{DR Pixel}$_\text{Trained}$\xspace}
\newcommand{\benchmark}{\textsc{DRPixelBench}\xspace}

\definecolor{okgreen}{RGB}{22,140,68}
\definecolor{nored}{RGB}{198,40,40}
\newcommand{\yes}{\textcolor{okgreen}{\ding{51}}}
\newcommand{\no}{\textcolor{nored}{\ding{55}}}

\usepackage{hyperref}

\usepackage[capitalise,noabbrev]{cleveref}

\crefformat{section}{\S#2#1#3}
\Crefformat{section}{\S#2#1#3}
\crefformat{subsection}{\S#2#1#3}
\Crefformat{subsection}{\S#2#1#3}
\crefmultiformat{section}{\S\S#2#1#3}{ and~#2#1#3}{, #2#1#3}{ and~#2#1#3}

\let\oldparagraph\paragraph
\renewcommand{\paragraph}[1]{\vspace{-.8em}
\oldparagraph{#1}
}

\usepackage{url}
\definecolor{darkblue}{rgb}{0, 0, 0.5}
\hypersetup{colorlinks=true, citecolor=darkblue, linkcolor=darkblue, urlcolor=darkblue}

\crefname{tcolorbox}{box}{boxes}
\Crefname{tcolorbox}{Box}{Boxes}

\newcommand{\huggingface}{\makebox[0pt][l]{\raisebox{-1.5pt}{\includegraphics[height=1.05em]{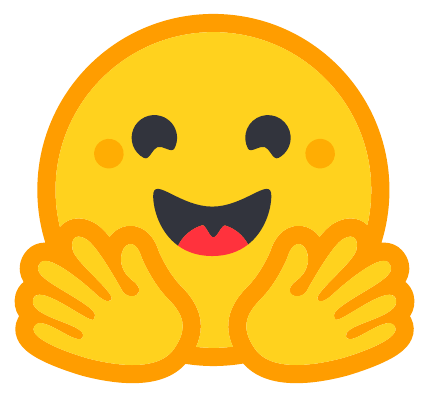}}}\xspace}
\newcommand{\github}{\makebox[0pt][l]{\raisebox{-1.5pt}{\includegraphics[height=1.05em]{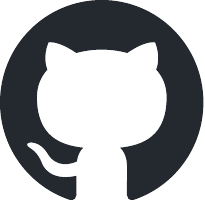}}}\xspace}
\newcommand{\titleLogo}{\includegraphics[height=2.3em]{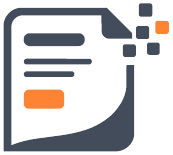}}

\definecolor{NYUPurple}{RGB}{134, 1, 175}
\definecolor{BerkeleyBlue}{RGB}{0, 50, 98}
\definecolor{RDColor}{RGB}{0, 122, 110}
\definecolor{TTICColor}{RGB}{176, 30, 45}
\newcommand{\NYUSH}{^{\hspace{.1em}\text{\textcolor{NYUPurple}{\textbf{N}}}}}
\newcommand{\UCB}{^{\hspace{.1em}\text{\textcolor{BerkeleyBlue}{\textbf{B}}}}}
\newcommand{\RD}{^{\hspace{.1em}\text{\textcolor{RDColor}{\textbf{R}}}}}
\newcommand{\TTIC}{^{\hspace{.1em}\text{\textcolor{TTICColor}{\textbf{T}}}}}
\newcommand{\NYUN}{^{\hspace{.1em}\text{\textcolor{NYUPurple}{\textbf{NY}}}}}

\newcommand{\affsep}{{}}

\title{%
\begin{tabular}{@{}m{2.6em}@{\hspace{0.5em}}m{\dimexpr\textwidth-3.1em\relax}@{}}
\titleLogo &
\raggedright \ours: Page-Native Deep Research with Vision-Language Models
\end{tabular}}

\author{%
Tiansheng Hu$\NYUSH$ \quad
Yichuan Wang$\UCB$ \quad
James V.\ Talwar$\RD$ \vspace{2pt}\\
\bf{%
Nicholas Tomlin$\TTIC\affsep\RD$ \quad
Chen Zhao$\NYUSH\affsep\NYUN$ \quad
Sewon Min$\UCB$%
}\vspace{3pt}\\
$\NYUSH$NYU Shanghai \quad
$\UCB$University of California, Berkeley \quad
$\RD$Rational Dynamics \vspace{1pt}\\
$\TTIC$Toyota Technological Institute at Chicago \quad
$\NYUN$New York University \vspace{4pt}\\
{\small
\github\ ~~~~~\textbf{Code:} \href{https://github.com/StarTrail-org/DRPixel}
{\path{StarTrail-org/DRPixel}} \quad
\huggingface\ ~~~~~\textbf{Data:} \href{https://huggingface.co/datasets/HughieHu/drpixel-bench}{\path{HughieHu/DRPixel-Bench}}}%
}

\iclrfinalcopy

\begin{document}
\addtocontents{toc}{\protect\setcounter{tocdepth}{-1}}
\maketitle
\lhead{Preprint}

\begin{abstract}

Complex documents such as scientific papers contain rich information in tables, figures, and equations; however, existing deep research (DR) agents typically reduce papers to OCR-extracted text. We instead ask whether vision-language models (VLMs) can enable \emph{page-native deep research}: directly retrieving and reasoning over rendered document page images. We introduce \ours, the first open-source end-to-end multimodal DR agent that uses rendered page images as its sole retrieval and reasoning unit, operating over a corpus of 1.06M pages. To evaluate this setting, we construct \benchmark, a benchmark on scientific papers requiring reasoning over text, figures, tables, and equations.
Surprisingly, page-native DR already works well without training, outperforming OCR-based baselines across state-of-the-art models like GPT-5.4, Gemini-3.5-Flash, and Qwen3.6-27B, while requiring fewer tool calls.

We next turn to DR training.
A key challenge is that teacher trajectories for agentic supervised finetuning are noisy, mixing critical reasoning and action steps with redundant or failed attempts.
To address this, we introduce \emph{counterfactual credit reweighting}, a fine-grained credit assignment method that weights individual reasoning and action steps by their counterfactual contribution to the final answer.
Our training improves retrieval recall by about 15 points and final accuracy by over 10 points on \benchmark, and likewise outperforms the untrained baseline on four existing benchmarks. On out-of-distribution ViDoRe-V3, \ours comes within $1.8$ points of an oracle that is given the gold pages directly. Our results suggest that the conventional \emph{OCR-then-retrieve} pipeline is not only unnecessary but suboptimal: directly retrieving and reasoning over visual document pages makes complex DR both more effective and more efficient.

\end{abstract}

\section{Introduction}

Deep research (DR) agents answer complex information-seeking questions by iteratively searching, reading, and reasoning over large collections of evidence~\citep{openai2025deepresearch,gemini2025deepresearch,li2025webthinker,tongyideepresearchteam2025tongyideepresearchtechnicalreport}. As DR expands beyond the open web into domains such as law, finance, and science, it increasingly operates over specialized document collections~\citep{li-etal-2025-legalagentbench, choi2026finagentbench, xie2024finben, chen2025medbrowsecomp, chen2025scienceagentbench}. These documents are inherently multimodal: critical evidence often lies not only in text, but also in tables, figures, charts, and equations.

Yet, such documents are typically reduced to plain text: PDFs are linearized into text, and retrieval and reasoning operate entirely in text space. This introduces two key limitations. First, linearization discards or distorts visual structure, making information in figures, tables, or charts difficult or impossible to recover~\citep{poznanski2025olmocr, lo2023semanticreaderprojectaugmenting}. Second, extraction and OCR errors propagate into downstream retrieval and reasoning~\citep{zhang2025ocrhindersragevaluating,wang2026pixelragwebscreenshotsbeat}.

This raises a natural question: \emph{Can deep research instead be page-native, retrieving and reasoning directly over rendered document pages?} In such an approach, each page remains an image: a visual retriever searches over rendered pages, and a vision-language model (VLM) reasons directly over the retrieved images.
Prior work has shown that such page-native retrieval can match or outperform text-based document retrieval~\citep{faysse2025colpali, moreira2026nemotroncolembedv2topperforming}, and has extended this approach to retrieval-augmented generation (RAG) tasks.
We study this approach in the deep research regime, where answering a query requires many rounds of retrieval and reasoning. To this end, we introduce \ours, the first open-source \emph{page-native} multimodal DR agent that iteratively searches and reasons directly over 1.06M rendered pages from scientific papers.

\begin{figure*}[t]
\centering
\includegraphics[width=\textwidth]{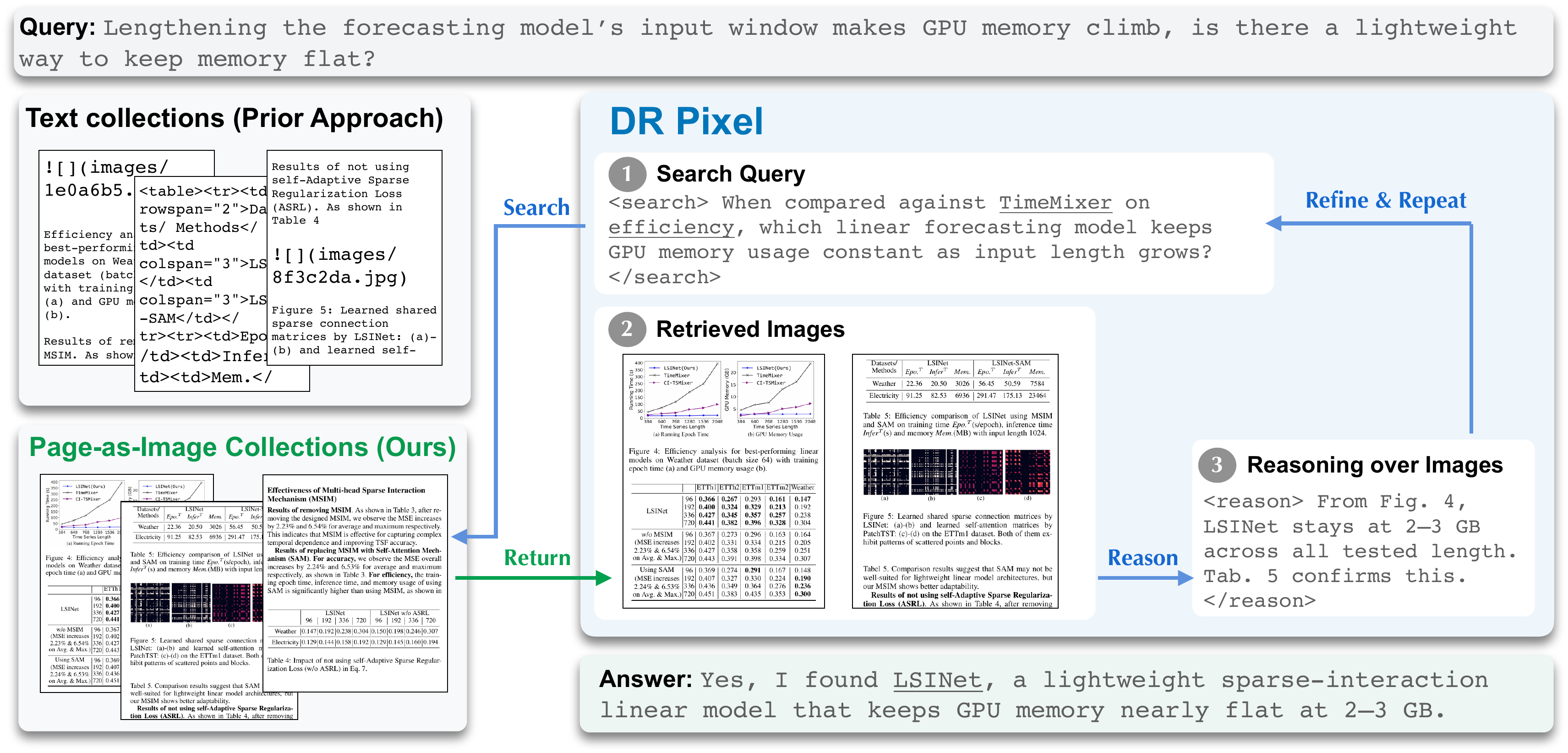}
\caption{Overview of \ours.
DR agents alternate between search, retrieval, and reasoning. Prior work operates in text space over OCR-extracted papers; \ours operates natively in visual space, with both retrieval and reasoning performed directly over rendered page images.
}
\label{fig:DR_Pixel}
\end{figure*}

The first challenge is evaluation:
existing DR benchmarks are either predominantly text-based~\citep{chen-etal-2026-browsecomp,xiong2026autoresearchbenchbenchmarkingaiagents,asai2024openscholarsynthesizingscientificliterature} or require relatively simple retrieval where a single search already achieves high recall~\citep{wang-etal-2025-vidorag,tanaka2023slidevqadatasetdocumentvisual}.
To bridge this gap, we introduce \benchmark, comprising $500$ questions over $1$M pages of scientific papers. It combines complex deep research queries that require long chains of retrieval and reasoning with the visual reasoning demands of scientific papers, such as interpreting experimental results in figures and tables.
Each question is grounded in a gold page and paired with a directly verifiable answer.

We then instantiate page-native deep research as an end-to-end VLM agent. \ours\ indexes rendered pages directly with a VLM embedding model~\citep{li2026qwen3vlembeddingqwen3vlrerankerunifiedframework}, feeds retrieved images to the VLM agent, and maintains them as visual evidence throughout the trajectory, without ever being converted into plain text. The agent iteratively formulates queries, retrieves images, and reasons over evidence accumulated across turns to generate the final answer.
Our evaluation on \benchmark reveals a consistent advantage of page-native DR over OCR-based pipelines, even without targeted training: page-native DR improves accuracy by $5.1$--$8.4$ points across GPT-5.4~\citep{gpt_5_4}, Gemini-3.5-Flash~\citep{gemini_3_5_flash}, and Qwen3.6-27B~\citep{qwen_3_6}, while using $20\%$--$30\%$ fewer retrieval calls than OCR-based pipelines.

Still, the open-weight Qwen3.6-27B lags behind closed models (40.1 vs. 51.9), motivating us to train \ours.
A key challenge is that demonstration traces for supervised fine-tuning (SFT) are often noisy, and are a mix of useful and unhelpful steps: some are critical to reaching the correct answers, while others are
 redundant or failed attempts.
To account for this, we introduce \emph{counterfactual credit reweighting}: a general recipe for DR training that weights each reasoning and action step by its contribution to the final answer.
For each step, we estimate its contribution by removing it from the trajectory and assigning higher training weight when its removal degrades answer quality.

Our training method significantly improves the performance: with Qwen3.6-27B, it improves accuracy by $12.0$ points ($30$\% relative) over the untrained model, and closes the gap to GPT-5.4. The gains also transfer beyond \benchmark: across three additional multimodal scientific literature benchmarks, our trained \ours\ remains competitive and surpasses GPT-5.4 in retrieval on two of them. Notably, on the out-of-distribution ViDoRe-V3 benchmark~\citep{loison-etal-2026-vidore}, the trained \ours comes within $1.8$ points of an oracle that is given the gold pages directly.

We summarize our key contributions as follows:
\begin{enumerate}[leftmargin=*, itemsep=2pt, topsep=2pt, parsep=0pt, label=\arabic*.]
    \item We introduce \ours, a page-native deep research agent that treats papers as collections of rendered page images and iteratively retrieves and reasons over them. Across GPT-5.4, Gemini-3.5, and Qwen3.6, \ours\ consistently outperforms OCR-based models that convert papers to text.

    \item We introduce a new DR training method, \emph{counterfactual credit reweighting}, which weights each reasoning and search step in the training data by its contribution to the gold answer.  It improves over the untrained model on all five benchmarks, and outperforms standard SFT.

    \item We release \benchmark, consisting of complex questions that require both visual document understanding (e.g., analyzing results from tables and figures) and long chains of retrieval and reasoning, providing a challenging testbed for future work on deep research.

\end{enumerate}

\section{Related Work}

\oldparagraph{Deep research (DR) agents.}
DR agents tackle complex, multi-step information-seeking tasks through iterative search and synthesis~\citep{yao2023react,li2025webthinker,li-etal-2025-search,shao2026dr}. Recent work extends the pipeline to multimodal deep research by adding tools such as image search, reverse image search or cross-modality retrieval~\citep{geng2026webwatcher,yao2026mmdeepresearch}. Most of them focus on open web, where text and images are retrieved separately through two different search tools~\citep{tao2026mmsearchplus,li2026mmbrowsecompcomprehensivebenchmarkmultimodal,jiang2024mmsearchbenchmarkingpotentiallarge}. We instead target a collection of documents (e.g., scientific papers), where answering complex questions often requires jointly interpreting text, tables, and figures on the same page, making separate text and image retrieval interfaces suboptimal.

\paragraph{Page-native document retrieval.}
With rapid progress with vision-language models (VLMs), bypassing OCR pipelines and directly processing rendered document images has emerged as a promising direction~\citep{kim2022ocrfreedocumentunderstandingtransformer,lee2023pix2structscreenshotparsingpretraining,blecher2024nougat,wei2026deepseekocr2visualcausal}.
Building on this, visual document retrieval retrieves rendered pages directly, established by ColPali and followed by increasingly capable visual retrievers~\citep{faysse2025colpali,gunther-etal-2025-jina,li2026qwen3vlembeddingqwen3vlrerankerunifiedframework}.
More recent work extends this approach to further generate answers~\citep{yu2025visrag,wang-etal-2025-vidorag,wang2026pixelragwebscreenshotsbeat} and add iterative retrieval~\citep{wang2025vragrl,shen2026visoragenticvisualretrievalaugmented}. However, these settings require relatively shallow search, e.g., on ViDoSeek~\citep{wang-etal-2025-vidorag} and SlideVQA~\citep{tanaka2023slidevqadatasetdocumentvisual}, a single retrieval step already recalls the gold page in the top five for $>90\%$ of questions~\citep{wang-etal-2025-vidorag,giouroukis2025whatsbestwayretrieve}. We instead study page-native deep research where complex queries require sustained search and reasoning over $10$--$20$ steps.

\paragraph{Agent training.}
Broadly, agents that generate long action trajectories are commonly trained with outcome-based reinforcement learning (RL)~\citep{jin2025searchr,zheng-etal-2025-deepresearcher,wang2025vragrl,shen2026visoragenticvisualretrievalaugmented}, sometimes with rubrics~\citep{shao2026dr}, or through supervised fine-tuning (SFT) on teacher-generated trajectories~\citep{chen2023fireactlanguageagentfinetuning,tongyideepresearchteam2025tongyideepresearchtechnicalreport}. Both provide limited credit assignment across a trajectory: outcome-based RL assigns a single reward to the rollout, and SFT imitates every teacher step regardless of its utility. Prior work has explored finer-grained supervision through process reward models that score each intermediate step~\citep{lightman2023letsverifystepstep} or by leveraging token- or step-level supervision coming from a teacher model conditioned on privileged context ~\citep{meng2026craftcounterfactualcreditassignment,wu2026outcomerewardssteplevelselfdistilled}, both during RL.
In this paper, we propose an alternative approach that retains the simplicity of SFT: we reweight each agentic step by its counterfactual contribution to the final answer, estimated by removing the step and measuring the resulting change in answer quality.

\section{DRPixelBench}
\label{sec: benchmark}

Existing benchmarks for complex information-seeking document collections largely fall into two categories: text-based DR benchmarks and multimodal document QA benchmarks (\cref{tab:benchmark-uniqueness}).
Text-based deep research benchmarks evaluate iterative search and synthesis over large corpora, but often rely on templated questions~\citep{chen-etal-2026-browsecomp} or long-form answers that require nondeterministic, rubric-based evaluation~\citep{du2026deepresearch,asai2024openscholarsynthesizingscientificliterature}; crucially, they are text-only.
Multimodal QA benchmarks instead require visual understanding from reports or scientific papers~\citep{ma2024mmlongbenchdoc,tanaka2023slidevqadatasetdocumentvisual,wang-etal-2025-vidorag,huang2026airqa,li-etal-2024-m3sciqa,shorten2026irpapersvisualdocumentbenchmark,loison-etal-2026-vidore}. However, their questions are often templated and require relatively shallow retrieval, with only one or a few steps needed to locate the relevant evidence.

\begin{wraptable}{r}{0.5\textwidth}
\centering\footnotesize
\setlength{\tabcolsep}{2pt}
\begin{tabular}{@{}lcccc@{}}
\toprule
Benchmark & Agentic & MM & Non-templ. & Verif. \\
\midrule
\multicolumn{5}{@{}l}{\textit{Deep research}} \\
BrowseComp-Plus    & \yes & \no  & \no  & \yes \\
DeepResearch Bench & \yes & \no  & \yes & \no  \\
ScholarQABench     & \yes & \no  & \yes & \no  \\
\addlinespace
\multicolumn{5}{@{}l}{\textit{Multimodal QA}} \\
AirQA (comp.)      & \yes & \yes & \no  & \yes \\
M3SciQA            & \yes & \yes & \no  & \yes \\
IRPAPERS           & \no  & \yes & \yes & \yes \\
ViDoSeek           & \no  & \yes & \no  & \yes \\
SlideVQA           & \no  & \yes & \no  & \yes \\
ViDoRe V3          & \no  & \yes & \no  & \yes \\
\midrule
\textbf{DRPixelBench} & \yes & \yes & \yes & \yes \\
\bottomrule
\end{tabular}
\caption{\benchmark is the only benchmark combining iterative search over a large corpus (\textbf{Agentic}), multimodal evidence (\textbf{MM}), non-templated questions (\textbf{Non-templ.}), and verifiable short answers (\textbf{Verif.}).}
\label{tab:benchmark-uniqueness}
\vspace{-2em}
\end{wraptable}

We evaluate on four existing benchmarks (IRPAPERS, M3SciQA, AirQA, ViDoRe-V3) but argue that page-native DR requires a new benchmark, meeting four key properties: (1) \emph{agentic search}, where relevant evidence typically requires multiple rounds of retrieval and reasoning; (2) \emph{multimodal evidence}, where answers depend on text together with figures, tables, or equations; (3) \emph{non-templated questions}, avoiding fixed schemas; and (4) \emph{short, verifiable answers}, allowing direct evaluation against gold answers without relying on imperfect rubrics.

\paragraph{\benchmark.}
To this end, we build \benchmark over scientific literature. Scientific papers are naturally multimodal, with important information distributed across text, tables, figures, and equations.

We collect papers from arXiv and the open-access proceedings of major conferences, and merge in the papers of AirQA~\citep{huang2026airqa}, M3SciQA~\citep{li-etal-2024-m3sciqa} and IRPAPERS~\citep{shorten2026irpapersvisualdocumentbenchmark} so that \benchmark and these three benchmarks share one corpus of $99{,}055$ papers.

Inspired by \citet{dasigi-etal-2021-dataset}, we generate each question from a paper's abstract rather than from its full text to keep it tied to the paper's main contribution. As an abstract reports its findings with few supporting details, we first enrich it. Using Gemini-3.5-Flash~\citep{gemini_3_5_flash}, we extract atomic facts from the full paper, such as a comparison in a table or a trend in a figure, and add them to the abstract. We then prompt the model to write a question from the enriched abstract alone, so that the question is driven by curiosity rather than by evidence already in view. We further ask the model to locate the supporting evidence in the full paper and produce a short-form answer. Depending on the form of this evidence, each question is labeled \emph{Text} or \emph{Visual}: a Text question is answered from the running text, while a Visual question requires a table, figure, or equation.
Each QA pair then undergoes four automatic checks to ensure that the question is plausible, the gold evidence is sufficient, non-textual evidence is necessary, and the answer cannot be recovered from parametric knowledge alone.
Finally, human experts review the question and rewrite it for naturalness.
The resulting \benchmark contains $500$ QA pairs.
More details, including the data sources, parsing details, filter criteria, annotation details, and data statistics, are reported in \cref{app:benchmark_detail}.

\section{DR Pixel Inference}

We introduce \ours: a deep research agent that uses rendered page images as both the retrieval unit and the model input. We describe our method (\cref{sec:inference_method}) and present results showing that \ours consistently outperforms OCR-based pipelines (\cref{sec: prelim_exp}).

\subsection{Inference Method}
\label{sec:inference_method}

Following \citet{tongyideepresearchteam2025tongyideepresearchtechnicalreport}, we formulate a DR agent as a ReAct loop~\citep{yao2023react}. At each turn $t$, the agent produces a thought and an action,
$(\tau_t, a_t) \sim \pi(\cdot \mid \mathcal{H}_{t-1}),$ where $\mathcal{H}_{t-1}$ denotes the history of previous thoughts, actions, and search results.
An action $a_t$ is either a search query or a final answer.
For a search action, a search tool $f$ returns search results $f(a_t, \mathcal{D}) \subset \mathcal{D}$ where $\mathcal{D}$ is a fixed corpus that a search tool interacts with. A final answer action terminates the trajectory. Thus, the complete trajectory that generates a final answer after $T$ tool calls is $[\mathcal{H}_T; \tau_{T+1}; a_{T+1}],$ where
\begin{equation}
\mathcal{H}_T=(\tau_1,a_1,f(a_1, \mathcal{D}),\ldots,\tau_T,a_T,f(a_T, \mathcal{D})).
\label{eq:react}
\end{equation}

\paragraph{Prior work.}
Prior work commonly parses documents into plain text, using document parsers or OCR, then operates the DR agent entirely in text space~\citep{chen-etal-2026-browsecomp,asai2024openscholarsynthesizingscientificliterature}. Formally, $\mathcal{D}$ is a collection of parsed text documents, $f$ is a text retrieval model (e.g., a dense retrieval model~\citep{karpukhin2020dense} that indexes documents or text chunks into fixed-size vectors), and $f(a_t, \mathcal{D})$ returns text chunks most relevant to query $a_t.$ The agent $\pi$ can be a text-only model.

However, a key limitation of this design is the lossy conversion from documents to text~\citep{niu2026mineru}.
Tables and equations may be parsed incorrectly; figures are often reduced to captions or omitted entirely; and OCR or parsing errors can corrupt even text content. Once discarded during parsing, this information cannot be recovered by downstream retrieval or reasoning.

Some prior work mitigates this limitation by adding a second, image-retrieval tool and allowing the agent to choose between text and image search~\citep{suri-etal-2025-visdom}. However, this is still limited because text and visual content are treated as separate spaces without preserving relationships between them, e.g., a figure and the surrounding text that explains it. Moreover, information loss or corruption during text parsing still remains.

\paragraph{\ours.}
Our DR agent, \ours, is an alternative instantiation of the DR agent in~\autoref{eq:react}: it preserves the ReAct loop but replaces the search tool $f$ and its corpus $\mathcal{D}$ to be image-based, as illustrated in~\autoref{fig:DR_Pixel}.
Offline, each paper is rendered page by page, so $\mathcal{D}$ becomes a collection of page images. Each image is encoded into a fixed-size vector using a VLM-based embedding model, e.g., Qwen3-VL-Embedding-8B. At inference time, $f$ maps a text query $a_t$ into the same embedding space and returns the top-$k$ page images.
These images are passed directly to the agent $\pi,$ which is itself a VLM. The agent reasons over those images and generates the next thought $\tau_{t+1}$ and action $a_{t+1},$ repeating this process until it produces a final answer.

\subsection{Results}
\begin{table}[t]
\centering
\setlength{\tabcolsep}{5pt}
\setlength{\fboxsep}{1pt}
\footnotesize
\begin{tabular}{l cc cc cc}
\toprule
\multirow{2}{*}{Retrieval Setup}
 & \multicolumn{2}{c}{GPT-5.4}
 & \multicolumn{2}{c}{Gemini-3.5-Flash}
 & \multicolumn{2}{c}{Qwen3.6-27B} \\
\cmidrule(lr){2-3}\cmidrule(lr){4-5}\cmidrule(lr){6-7}
 & Acc ($\uparrow$) & \# Calls ($\downarrow$) & Acc ($\uparrow$) & \# Calls ($\downarrow$) & Acc ($\uparrow$) & \# Calls ($\downarrow$)\\
\midrule
\multicolumn{7}{@{}l}{\textit{OCR-based pipelines}} \\
\;Text
 & 43.5 & 12.4
 & 40.0 & 17.5
 & 33.2 & 11.0 \\
\;\;+ Figure (\colorbox{text_high}{txt}+\colorbox{figure_high}{fig})
 & 46.3 & \colorbox{text_high}{10.8}+\colorbox{figure_high}{3.0}
 & 41.3 & \colorbox{text_high}{16.7}+\colorbox{figure_high}{1.0}
 & 34.9 & \colorbox{text_high}{10.9}+\colorbox{figure_high}{0.6} \\
\midrule
\ours
 & \textbf{51.9} & \textbf{9.2}
 & \textbf{46.4} & \textbf{13.7}
 & \textbf{40.1} & \textbf{8.8} \\
\bottomrule
\end{tabular}
\caption{\ours versus OCR-based retrieval pipelines, all with no training. OCR-based pipelines retrieve OCR text chunks (\emph{Text}), optionally with a separate search on the figure corpus (\emph{+ Figure}).
\ours retrieves and reads whole pages as images. \textbf{Acc} is answer accuracy. \textbf{\# Calls} is the number of tool calls per question, split into \colorbox{text_high}{text} and \colorbox{figure_high}{figure} calls in the text-and-figure setting.
}
\label{tab:page_better}
\end{table}

\label{sec: prelim_exp}
We now compare \ours against two OCR-based baselines---text-only and text+figure from \cref{sec:inference_method}---on \benchmark.
\paragraph{Setup.}
We evaluate GPT-5.4~\citep{gpt_5_4}, Gemini-3.5-Flash~\citep{gemini_3_5_flash}, and Qwen3.6-27B~\citep{qwen_3_6} with \texttt{reasoning\_effort=high} for the first two and \texttt{enable\_thinking=True} for Qwen3.6-27B. All models run in the same ReAct loop with at most $T=20$ searches, and all observations remain in context. We report accuracy judged by GPT-5.4-mini~\citep{gpt_5_4} under a strict prompt, taking a majority vote over three independent judgments.

From the same $99{,}055$ papers, we build three corpora, one for each setting.
The text-only baseline parses each PDF into text, preserving tables as HTML and equations as \LaTeX, and retrieves $k=5$ chunks of 256 words, with Qwen3-Embedding-8B~\citep{zhang2025qwen3embeddingadvancingtext}.
The text+figure baseline additionally retrieves $k=3$ extracted figures when the image-search tool is invoked, using Qwen3-VL-Embedding-8B~\citep{li2026qwen3vlembeddingqwen3vlrerankerunifiedframework}.
\ours, instead, searches $1{,}062{,}153$ page images rendered directly from the PDFs, returning $k=3$ pages, also using Qwen3-VL-Embedding-8B.
We choose $k$ for each setting to approximately match the agent's retrieved token budget (\autoref{tab:retrieval-budget} for details).

\begin{figure}[t]
\centering
\begin{minipage}[t]{0.48\textwidth}
  \vspace{0pt}
  \centering
   \includegraphics[width=\linewidth]{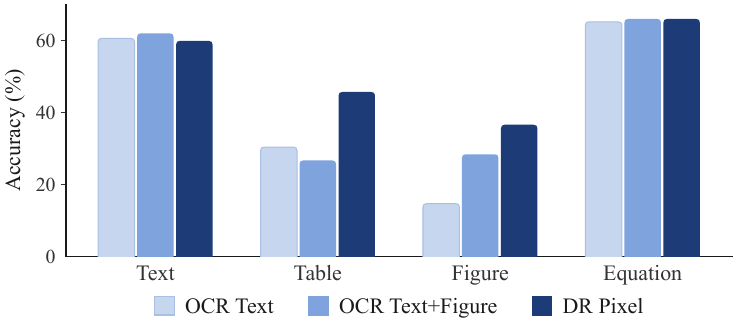}
  \caption{GPT-5.4 accuracy by evidence source under the two OCR-based settings and \ours. \ours consistently outperforms OCR-based pipelines on questions that require tabular and figure evidence.
  }
  \label{fig:modality_gpt_settings}
\end{minipage}\hfill
\begin{minipage}[t]{0.48\textwidth}
  \vspace{0pt}
  \captionsetup{type=table}
  \centering\footnotesize
  \setlength{\tabcolsep}{1pt}
  \begin{tabular}{@{}lccr@{}}
  \toprule
   & GPT-5.4 & Qwen3.6-27B & $\Delta$ \\
  \midrule
  Accuracy                   & 51.9 & 40.1 & $-11.8$ \\
  \midrule
  \multicolumn{4}{@{}l@{}}{\textit{Retrieval}} \\
  Page Hit                   & 45.7 & 36.3 & $-9.4$ \\
  Repeated Queries           & 0.0  & 5.3  & $+5.3$ \\
  \midrule
  \multicolumn{4}{@{}l@{}}{\textit{Reading}} \\
  Acc $\mid$ Page Hit        & 71.2 & 63.9 & $-7.3$ \\
  \;\;Hit on Last Search     & 74.4 & 69.3 & $-5.1$ \\
  \;\;Hit on Earlier Search  & 69.7 & 57.4 & $-12.3$ \\
  \bottomrule
  \end{tabular}
  \captionof{table}{Decomposing gap between GPT-5.4 and Qwen3.6-27B. Qwen3.6-27B retrieves the gold page less often and uses it less well once found.}
  \label{tab:gap-decomposition}
\end{minipage}
\end{figure}

\paragraph{Page-native deep research outperforms OCR-based pipelines.}
As shown in~\autoref{tab:page_better}, \ours consistently outperforms both OCR-based settings across all three models, improving accuracy over the text-and-figure pipeline by $5.6$, $5.1$, and $5.2$ points for GPT-5.4, Gemini-3.5-Flash, and Qwen3.6-27B respectively. Adding a figure-search tool yields only marginal gains, as agents rarely invoke it despite explicit instructions in the prompt. \ours is also more efficient, reducing retrieval calls by $20\%$ to $30\%$. We further break down GPT-5.4's accuracy by evidence source in~\autoref{fig:modality_gpt_settings}. \ours achieves the highest accuracy on questions requiring tabular and figure evidences, and remains on par on those grounded in text or equations.

\paragraph{Open-weight models lag in both retrieval and reasoning.}
Although Qwen3.6-27B issues a similar number of retrieval calls to GPT-5.4, it trails by $11.8$ points in accuracy under \ours setting. As shown in~\autoref{tab:gap-decomposition}, we decompose this gap into retrieval and reasoning. In retrieval, Qwen3.6-27B retrieves the gold page $9.4$ points less often, and its searches are less productive with $5.3\%$ of its queries duplicate an earlier one and $21.6\%$ of its trajectories contain at least one duplicate. In reasoning, Qwen3.6-27B struggles to use pages retrieved earlier in the trajectory. When the gold page comes from the latest search, it trails GPT-5.4 by only $5.1$ points, but the gap widens to $12.3$ when the page was retrieved in an earlier search.
This motivates training an open-weight model, as described in \cref{sec:training}.

\section{DR Pixel Training}\label{sec:training}

Having shown that page-native DR improves performance even without training, we next train \ours to further improve its capabilities. We denote the untrained model as \oursbase and the trained model as \ourstrained. We first describe the training method (\cref{sec:training_method}), then present the main results (\cref{sec:main_results}) and ablations (\cref{sec: ablation}).

\subsection{Training Method}
\label{sec:training_method}

We train \ours with supervised fine-tuning (SFT). We sample trajectories from GPT-5.4 on training questions generated using the same pipeline as \S\ref{sec: benchmark}, but without human rewriting or verification; and use them as training data.

However, we find standard SFT has a problem because the trajectories in the training data often contain repeated or low-value searches that do not contribute to the final answer.
Naively imitating these unhelpful steps encourages excessive searching and inefficient use of the retrieval budget. Inspired by process rewards, which assign credit to individual intermediate steps~\citep{lightman2024lets,setlur2025rewarding,zhou2024archer,zhou2025sweetrltrainingmultiturnllm}, we instead weight each step by its contribution to the final answer. Specifically, our \emph{counterfactual credit reweighting} removes each step from a fixed teacher trajectory and measures the resulting drop in the gold-answer likelihood under the frozen base model, assigning higher weight to steps that are more consequential.

\begin{figure*}[t]
\centering
\includegraphics[width=\textwidth]{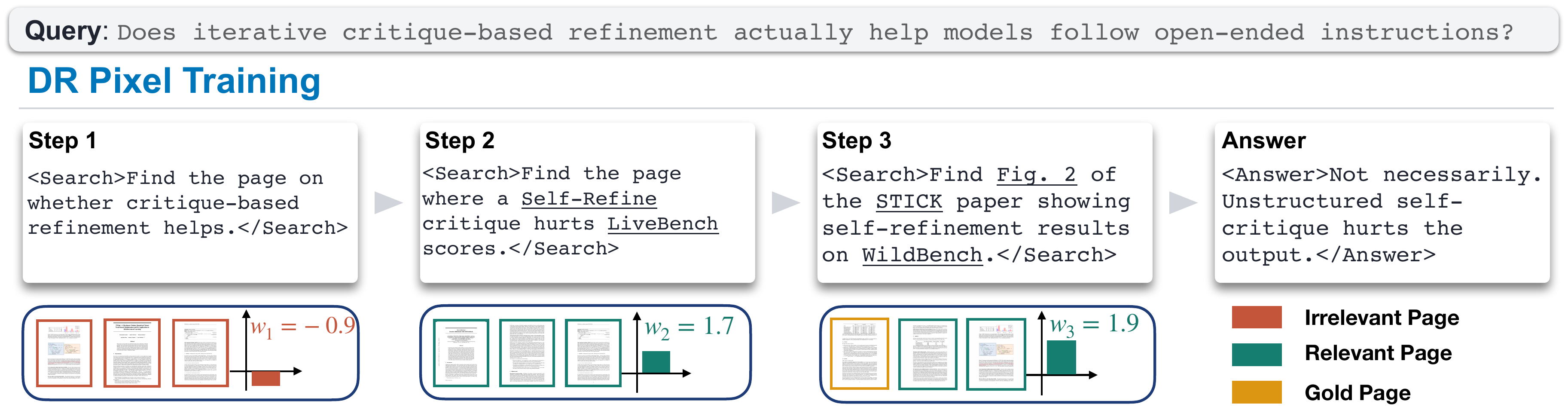}
\caption{
An illustration of \emph{counterfactual credit reweighting} (\S\ref{sec:training_method}).
The first search returns three off-topic pages and lowers the gold score, so the step has a negative score and weight $0$. The second retrieves the gold paper, though not the exact gold page. This increases the gold score and receives positive weight. The third retrieves the exact evidence pages and likewise receives positive weight.
}
\label{fig:credit_qualitative}
\end{figure*}

\paragraph{Counterfactual credit.}

Formally, let $y^\star$ denote the gold answer, $\mathcal{H}_i$ the history after $i$ search calls as defined in \autoref{eq:react}, and $\pi_0$ the frozen base policy. For any history $\mathcal{H}$, we measure how strongly it supports the gold answer by the log-likelihood that $\pi_0$ assigns to $y^\star$:
\begin{equation}
\ell(\mathcal{H}) = \log \pi_0\!\left(y^\star \mid \mathcal{H}\right).
\end{equation}
We assign credit separately to reasoning and search steps. For a reasoning step $\tau_i$, we measure how much removing it from $\mathcal{H}_T$ decreases support for the gold answer. For a search step $a_i$, we measure how much the search results $f(a_i, \mathcal{D})$ increase this support relative to the history before the search:
\begin{equation}
c^{\tau}_i = \ell(\mathcal{H}_T)-\ell\!\left(\mathcal{H}_T \setminus \tau_i\right),
\qquad
c^{a}_i = \ell(\mathcal{H}_{i})-\ell(\mathcal{H}_{i-1}).
\label{eq:cf}
\end{equation}
Thus, $c_i^\tau$ captures the contribution of a reasoning step to organizing the accumulated evidence, while $c_i^a$ captures the value of the new evidence returned by a search.

\paragraph{Weight assignment and normalization.}

Because raw step credits are unbounded, we convert them into rank-based weights for each step type $u\in\{\tau,a\}$. Specifically, steps with non-positive credit receive weight $0$. For every positive credit, if $c_i^u$ is the $k$-th smallest among $n$ surviving steps of type $u$, we define $\mathrm{rank}(c_i^u)=k/n$. We then multiply by $2$ so the average positive weight remains near $1$, matching the scale of standard SFT. The resulting weight $w_i^u$ is:
\begin{equation}
w^{u}_i=
\begin{cases}
0, & c^{u}_i\le 0,\\[2pt]
2\cdot\mathrm{rank}\!\left(c^{u}_i\right), & \text{otherwise.}
\end{cases}
\label{eq:w}
\end{equation}
As each reasoning step directly produces its following search query, we further couple their weights. For each reasoning step, we define its likelihood gain as the increase in log-likelihood of its corresponding action relative to an empty reasoning block. We mask a reasoning step if its corresponding action was previously masked or its likelihood gain is lower than the median over all reasoning steps in the training data. A reasoning step that passes both checks keeps its weight $w_i^\tau$. If such a reasoning step was previously masked due to negative counterfactual credit
with $w_i^\tau=0$, it is unmasked and assigned $\bar{w}$, the median positive reasoning weight. We denote the resulting weight by $\tilde{w}_i^\tau$.

The final training objective is a token-level weighted loss, with a constant weight $\lambda$ on the final-answer tokens:
\begin{equation}
\mathcal{L}(\theta)=-\sum_{i=1}^{T-1}\!\left[
\tilde{w}^{\tau}_i\!\!\sum_{t\in\tau_i}\!\!\log\pi_\theta(x_t\mid x_{<t})
+w^{a}_i\!\!\sum_{t\in a_i}\!\!\log\pi_\theta(x_t\mid x_{<t})
\right]
-\,\lambda\!\!\sum_{t\in a_T}\!\!\log\pi_\theta(x_t\mid x_{<t}).
\end{equation}

\paragraph{Training details.}
We fine-tune Qwen3.6-27B with LoRA on about $2{,}500$ teacher trajectories from GPT-5.4 for one epoch, using a learning rate of $1\mathrm{e}{-4}$, an effective batch size of $16$, and answer weight $\lambda=1.5$, on four H200 GPUs.
More details are provided in \cref{app:training_setup}.

\subsection{Results}
\label{sec:main_results}

\oldparagraph{Setup.}
We evaluate on \benchmark and three external multimodal scientific literature benchmarks: IRPAPERS~\citep{shorten2026irpapersvisualdocumentbenchmark}, M3SciQA~\citep{li-etal-2024-m3sciqa}, and AirQA~\citep{huang2026airqa}. All four share a unified corpus, as we form the corpus as a union of corpora from each benchmark (\cref{sec: benchmark}).
To test out-of-distribution generalization, we further evaluate on ViDoRe-V3~\citep{loison-etal-2026-vidore} over its own document corpus, with two search tools: page retrieval as in \ours, and text retrieval as in the text-only setting of \cref{sec: prelim_exp}, where each page is parsed into text and retrieved with Qwen3-Embedding-8B.  We compare \ourstrained with GPT-5.4, Gemini-3.5-Flash, the untrained \oursbase, and vanilla SFT, which is trained on the same teacher trajectories with every token weighted equally. All agents follow the configuration in~\cref{sec: prelim_exp}. \begin{wraptable}{r}{0.5\textwidth}
\centering
\footnotesize\setlength{\tabcolsep}{3pt}
\begin{tabular}{@{}lcccc@{}}
\toprule
Model & Tool/q & PgHit & Acc$|$Pg & Acc \\
\midrule
\multicolumn{5}{@{}l}{\textit{Open-weight (Qwen3.6-27B)}} \\
\oursbase             & 8.8  & 36.3 & 63.9 & 40.1 \\
\quad + vanilla SFT   & 14.0 & \textbf{53.6} & 67.2 & 49.0 \\
\ourstrained          & 8.9  & \underline{52.0} & 67.7 & \textbf{52.1} \\
\midrule
\multicolumn{5}{@{}l}{\textit{Proprietary}} \\
GPT-5.4          & 9.2  & 45.7 & \underline{71.2} & \underline{51.9} \\
Gemini-3.5-Flash & 13.7 & 35.0 & \textbf{75.7} & 46.4 \\
\bottomrule
\end{tabular}
\caption{Results on \benchmark. \textbf{Tool/q} is searches per question. \textbf{PgHit} is the percentage of questions that retrieves the gold page, and \textbf{Acc$\mid$Pg} is accuracy restricted to those questions. Results are averaged over three runs.
\ourstrained matches GPT-5.4 in accuracy, retrieving the gold pages more often while using fewer searches.
}
\label{tab:in_distribution_result_ours}
\end{wraptable}

\paragraph{\ourstrained matches proprietary models.}
As shown in~\autoref{tab:in_distribution_result_ours}, training \ours significantly improves performance on \benchmark, increasing accuracy from $40.1$ to $52.1$ and even slightly surpassing GPT-5.4. Our trajectory analysis shows that the gains are concentrated in retrieval: \ourstrained retrieves the gold page $6.3$ points more often than GPT-5.4. This suggests that post-training better aligns query generation with the retriever's embedding space, therefore producing queries that are more effective for visual retrieval.

\paragraph{Counterfactual credit reweighting makes search more efficient.}
\autoref{tab:in_distribution_result_ours} shows that our counterfactual credit reweighting consistently outperforms vanilla SFT, primarily by suppressing the ineffective searches from teacher trajectories. The vanilla SFT achieves a slightly higher page-hit rate than \ourstrained, but requires $57\%$ more searches. The key difference is search efficiency: vanilla SFT exhausts the $20$-call budget on $35.8\%$ of questions, compared with only $9.8\%$ for \ourstrained. This is even lower than GPT-5.4 ($17.8\%$) and the untrained backbone ($19.2\%$).

\begin{table*}[t]\centering
\footnotesize\setlength{\tabcolsep}{2pt}
\begin{tabular}{@{}lcccccccccccc@{}}
\toprule
& \multicolumn{4}{c}{IRPAPERS} & \multicolumn{4}{c}{M3SciQA} & \multicolumn{4}{c}{AirQA} \\
\cmidrule(lr){2-5}\cmidrule(lr){6-9}\cmidrule(lr){10-13}
Model & T/q & DocHit & Acc$\mid$Doc & Acc & T/q & DocHit & Acc$\mid$Doc & Acc & T/q & DocHit & Acc$\mid$Doc & Acc \\
\midrule
\multicolumn{13}{@{}l@{}}{\textit{Open-weight (Qwen3.6-27B)}} \\
\oursbase    & 5.6 & 89.4 & 68.8 & 65.4 & 7.8 & 75.3 & 63.4 & 49.8 & 7.9 & 84.3 & 46.0 & 42.7 \\
\ourstrained  & 5.0 & \textbf{95.6} & \textbf{82.0} & \textbf{79.4} & 9.4 & \textbf{90.3} & \underline{69.6} & \underline{64.2} & 7.0 & \textbf{89.7} & \underline{55.6} & \underline{52.3} \\
\midrule
\multicolumn{13}{@{}l@{}}{\textit{Proprietary}} \\
GPT-5.4             & 5.4 & 92.2 & \underline{80.1} & \underline{77.8} & 7.7 & \textbf{90.3} & \textbf{73.3} & \textbf{67.5} & 6.2 & \underline{87.9} & \textbf{59.5} & \textbf{56.6} \\
Gemini-3.5-Flash    & 8.4 & \underline{93.3} & 70.1 & 67.6 & 12.4 & \underline{87.2} & 68.3 & 60.8 & 11.1 & 87.5 & 50.0 & 47.0 \\
\bottomrule
\end{tabular}
\caption{Results on external multimodal scientific-QA benchmarks. \textbf{T/q} is searches per question. \textbf{DocHit} is the percentage of questions for which the gold document is retrieved, and \textbf{Acc$\mid$Doc} is accuracy restricted to those questions. \ourstrained improves over \oursbase on all three benchmarks and adapts search effort dynamically.
}
\label{tab:in_distribution_result_three}
\end{table*}

\paragraph{\ourstrained generalizes to external multimodal QA benchmarks.}
As shown in~\autoref{tab:in_distribution_result_three}, the behavior learned on \benchmark transfers to IRPAPERS, M3SciQA, and AirQA. \ourstrained improves over the untrained backbone on all three, with accuracy gains ranging from $9.6$ to $14.4$ points. It also adapts its search effort to retrieval difficulty variations: compared with the backbone, \ourstrained searches less on IRPAPERS and AirQA but more on M3SciQA, where the backbone has the largest document-retrieval gap relative to GPT-5.4 (15.0 points).%

\begin{table*}[t]\centering\footnotesize\setlength{\tabcolsep}{3pt}
\begin{tabular}{@{}l l l ccccc ccc c@{}}
\toprule
& & & \multicolumn{5}{c}{English} & \multicolumn{3}{c}{French} & \\
\cmidrule(lr){4-8}\cmidrule(lr){9-11}
Retriever & Mod. & Generator
& C.S. & Fin. & Phar. & H.R. & Ind. & Phys. & Ener. & Fin. & Global \\
\midrule
\multirow{2}{*}{Oracle$^{\ast}$}
 & Text   & \multirow{2}{*}{Gemini 3 Pro}
 & 80.9 & 70.2 & 71.4 & 72.3 & 66.4 & 71.2 & 69.2 & 62.8 & 70.6 \\
 & Page  & & 86.5 & 70.6 & 76.1 & 71.1 & 68.2 & 74.5 & 69.8 & 64.1 & 72.6 \\
\midrule
\multirow{2}{*}{Qwen3-Emb-8B}
 & \multirow{2}{*}{Text} & \oursbase
 & 86.5 & 64.7 & 60.7 & 63.8 & 53.7 & 77.4 & 66.6 & 51.9 & 65.7 \\
 & & \ourstrained
 & \underline{88.8} & \textbf{71.2} & 67.3 & \underline{65.4} & \textbf{66.1} & \underline{77.5} & \underline{69.5} & \underline{52.8} & \underline{69.8} \\
\midrule
\multirow{2}{*}{Qwen3-VL-Emb-8B}
 & \multirow{2}{*}{Page} & \oursbase
 & 87.9 & 64.4 & \underline{67.9} & 64.2 & 59.7 & 76.2 & 62.7 & 50.9 & 66.7 \\
 & & \ourstrained
 & \textbf{89.3} & \underline{67.0} & \textbf{73.1} & \textbf{67.0} & \underline{65.4} & \textbf{79.1} & \textbf{70.1} & \textbf{55.3} & \textbf{70.8} \\
\bottomrule
\end{tabular}
\caption{End-to-end accuracy on ViDoRe~V3. Oracle$^{\ast}$ is ViDoRe~V3's published upper bound. \ourstrained improves over \oursbase with both page and text retrieval on documents outside its training domain, so the training generalizes well.
}
\label{tab:vidore-v3}
\end{table*}

\paragraph{\ourstrained further generalizes across domains and retrieval modalities.}
ViDoRe-V3 provides a strong out-of-distribution test, covering non-scientific documents from domains such as finance and human resources. As shown in~\autoref{tab:vidore-v3}, post-training improves both page-based and text-based retrieval by $4.1$ points, despite using only page-native teacher trajectories, suggesting that the learned search behavior transfers across retrieval modalities. The two settings stay close because only $1.7\%$ of ViDoRe-V3 questions genuinely require visual content to answer. We further compare against oracle retrieval settings, where the relevant document or page is provided directly. With page retrieval, \ourstrained matches the Text Oracle and remains competitive with the Page Oracle.

\subsection{Ablation}
We next ablate on a couple of key factors: corpus size, training data modalities and page resolutions.

\label{sec: ablation}
\begin{figure}[t]
\centering
\begin{minipage}[t]{0.58\textwidth}
  \vspace{0pt}
  \centering
  \includegraphics[width=\linewidth]{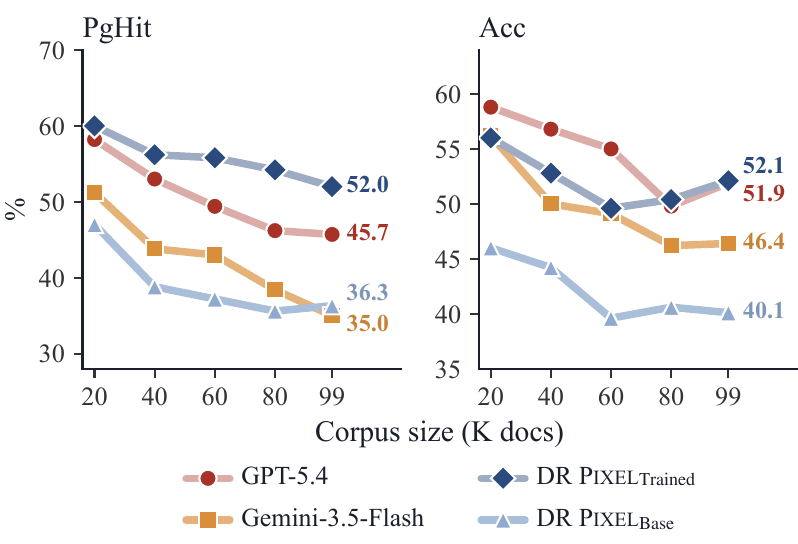}
  \caption{Effect of corpus size on retrieval and final answer accuracy. Each panel varies the number of indexed documents from $20{,}000$ to $99{,}055$. \ourstrained{} is the most robust to corpus growth, with the slowest degradation in retrieval performance.}
  \label{fig:corpus_scaling}
\end{minipage}\hfill
\begin{minipage}[t]{0.38\textwidth}
  \vspace{0pt}
  \centering
  \includegraphics[width=\linewidth]{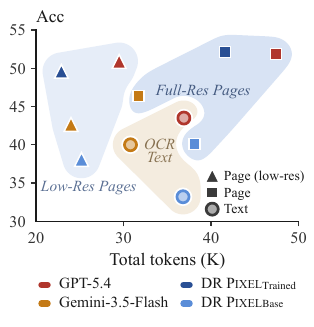}
  \caption{Accuracy against total tokens as page resolution varies. Lowering resolution cuts tokens by 24\% to 45\% for a 1 to 4 point drop, and page retrieval stays ahead of OCR text.}
  \label{fig:accuracy_efficiency}
\end{minipage}
\end{figure}

\begin{figure}[t]
\centering
\begin{minipage}[c]{0.47\textwidth}
  \centering
  \small
  \setlength{\tabcolsep}{3pt}
  \begin{tabular}{@{}lcccc@{}}
  \toprule
  Setup & Tool/q & PgHit & Acc$|$Pg & Acc \\
  \midrule
  \oursbase        & 8.8  & 36.3 & 63.9 & 40.1 \\
  + Text           & 5.0  & 47.4 & 60.3 & 46.0 \\
  + Visual, 2,000  & 10.1 & \textbf{52.2} & \textbf{65.5} & \textbf{50.2} \\
  \bottomrule
  \end{tabular}
\end{minipage}\hfill
\begin{minipage}[c]{0.5\textwidth}
  \centering
  \includegraphics[width=\linewidth]{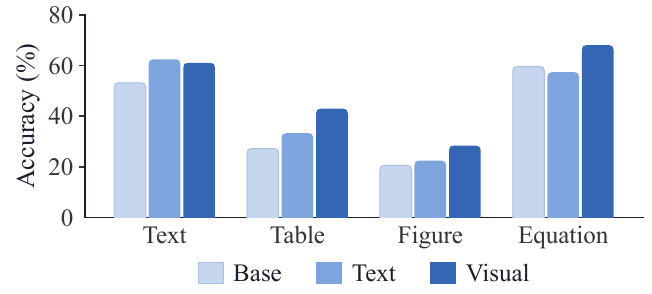}
\end{minipage}

\begin{minipage}[t]{0.47\textwidth}
  \captionof{table}{Effect of trajectory type on the distilled policy. Training only on Text trajectories makes the model search less, so it explores less and reaches a lower end-to-end accuracy.
  }
  \label{tab:setab-modality}
\end{minipage}\hfill
\begin{minipage}[t]{0.5\textwidth}
  \caption{Each training mixture improves the question type it matches, with Text trajectories helping text questions and visual trajectories helping multimodal ones.}
  \label{fig:modality_setab}
\end{minipage}
\end{figure}

\paragraph{\ourstrained is robust to corpus size.}
We scale the corpus from $20{,}000$ to $99{,}000$ papers while keeping the questions and gold documents fixed. As shown in~\autoref{fig:corpus_scaling}, page-hit rate declines for all models as the corpus grows, but \ourstrained degrades more slowly, dropping by only $8.0$ points compared with $12.5$ for GPT-5.4. As a result, its retrieval advantage over GPT-5.4 widens from $1.8$ points at $20{,}000$ papers to $6.3$ points at $99{,}000$, leading to final answer accuracy improvement.

\paragraph{Downgrading page resolution buys efficiency at a modest cost.}

In page-native deep research, page resolution determines the trade-off between visual fidelity and inference cost. Since processing high-resolution page images is expensive, we cap each rendered page at $1.05$ megapixels, roughly $50\%$ of its native resolution. As shown in~\autoref{fig:accuracy_efficiency}, this reduces input tokens by $30\%$ to $48\%$ across models, while accuracy drops by only $1.0$ to $3.8$ points. Output tokens rise slightly at lower resolution, but as they account for less than $25\%$ of all tokens, the overall savings largely remain. \footnote{We count each token once, as in the final context. Since every turn re-reads the full context in the ReAct loop, cumulative input is much larger, so output falls below $4\%$ of all tokens.} Even at this lower resolution, page retrieval outperforms OCR-based text retrieval by $2.7$ to $7.4$ points for every model evaluated while using $22\%$ to $33\%$ fewer input tokens. The breakdown into input and output tokens is shown in~\autoref{fig:accuracy_efficiency_split} in the appendix.

\paragraph{Fine-tuning on \emph{Visual} questions is more effective.}
The $2{,}500$ trajectories used to fine-tune \ourstrained contain only \emph{Visual} questions, as defined in~\cref{sec: benchmark}. To justify this choice, we generate \emph{Text} questions with the same pipeline, collect their trajectories as described in~\cref{app:training_setup}, and fine-tune the same base model on either $2{,}000$ \emph{Text} or $2{,}000$ \emph{Visual} trajectories, keeping all other settings fixed. As shown in~\cref{tab:setab-modality}, fine-tuning on Text questions causes the agent to stop earlier and retrieve the gold page less often, yielding $46.0$ accuracy versus $50.2$ with Visual fine-tuning. Our further analysis indicates that visual fine-tuning also transfers better: it remains within $1.4$ points of Text fine-tuning on \emph{Text} questions, while Text fine-tuning trails by $6$--$11$ points on \emph{Visual} questions.

\section{Conclusion}

Deep research over complex documents typically begins with OCR. \ours instead performs retrieval and reasoning directly over rendered pages. We study this setting with \benchmark, containing $500$ verifiable questions over $1.06$M pages of scientific literature, and show that page-native deep research consistently outperforms OCR-based pipelines across all tested models while using fewer tool calls, even without task-specific training. To close the remaining gap for open-weight models, we introduce \emph{counterfactual credit reweighting}, which weights teacher steps by their contribution to the gold answer and brings a $27$B model to GPT-5.4-level accuracy with lower search cost. These gains transfer to external benchmarks and non-scientific documents. More broadly, page-native deep research requires only document rendering rather than format-specific parsing, making it particularly attractive for private, layout-rich corpora such as financial or law documents where maintaining OCR and parsing pipelines is costly and brittle.

\section*{Acknowledgement}
We would like to thank Prasann Singhal, Ryan Wang, Rulin Shao, Zijian Chen and the SM group members for valuable discussions and feedback.

This work was supported in part by the Gemini Academic Program and by gifts from Rational Dynamics and Coefficient Giving. This work was also supported in part through the NYU IT High Performance Computing resources, services, and staff expertise.

\section*{AI Usage Statement}
LLMs are part of our pipeline: Gemini-3.5-Flash generates and filters questions (\cref{sec: benchmark}), with every benchmark question reviewed by human. GPT-5.4 produces training trajectories, and GPT-5.4-mini judges answers. We also used LLM assistants for literature search, writing polish, and coding. All research ideas, experimental design, and analysis are the authors' own.

\newpage
\bibliography{iclr2027_conference}
\bibliographystyle{iclr2027_conference}

\appendix

\newpage
\appendix
\addtocontents{toc}{\protect\setcounter{tocdepth}{3}}
\renewcommand{\contentsname}{\large Appendix Contents}
\hypersetup{linkcolor=black}
\tableofcontents
\newpage
\section{\benchmark Detail}
\label{app:benchmark_detail}
\subsection{OCR Parsing Detail}
All three corpora take the same PDFs as input and differ only in the retrieval unit they produce. We parse every PDF with MinerU-2.5-Pro~\citep{niu2026mineru} on A100 GPUs into typed blocks, from which both OCR-based corpora are built. The text corpus linearises the blocks in reading order, keeping equations as \LaTeX{} and tables as HTML with their captions and representing a figure by its caption alone. Its unit is a $256$-word chunk prefixed with the paper title. The figure corpus takes the cropped figure and chart images as its unit. The page-native corpus skips parsing entirely and renders each page at $150$ dpi, so its unit is a full page image.
\subsection{\benchmark Statistics Breakdown}
\begin{table}[htbp]
\centering
\small
\setlength{\tabcolsep}{3pt}
\begin{tabular}{@{}l@{\hspace{5pt}}rrrrr@{}}
\toprule
\textbf{Property} & \textbf{Text} & \textbf{Table} & \textbf{Figure} & \textbf{Equation} & \textbf{All} \\
\midrule
\multicolumn{6}{@{}l}{\textit{Questions}} \\
\quad Query Num   & 216  & 135  & 102  & 47   & 500  \\
\quad Query Len.  & 35.0 & 34.4 & 32.7 & 37.3 & 34.6 \\
\quad Answer Len. & 39.8 & 50.7 & 44.4 & 37.5 & 43.5 \\
\quad Gold Page   & 1.00 & 1.00 & 1.00 & 1.00 & 1.00 \\
\midrule
\multicolumn{6}{@{}l}{\textit{Corpus}} \\
\quad Papers      & \multicolumn{5}{r@{}}{99{,}055} \\
\quad Pages       & \multicolumn{5}{r@{}}{1{,}062{,}153} \\
\bottomrule
\end{tabular}
\caption{\benchmark statistics. Questions are grouped by the modality of the evidence they are
grounded in, and lengths are in tokens. Every question is grounded to exactly one page.}
\label{tab:dataset_statistics}
\end{table}

\subsection{\benchmark Verification Detail}
In~\autoref{sec: benchmark}, we present the verification we take to filter out low quality questions. Here, we present the total number of filters we apply.
\begin{table}[H]
\centering
\small
\setlength{\tabcolsep}{6pt}
\renewcommand{\arraystretch}{1.15}
\begin{tabular}{@{}lp{0.62\textwidth}r@{}}
\toprule
Filter & Criterion & Remaining \\
\midrule
\multicolumn{2}{@{}l}{Drafted questions} & $100.00\%$ \\
\midrule
Coined-name removal & The paper's own product or benchmark name is replaced by a generic class; questions where it cannot be removed are discarded. & $99.07\%$ \\
Self-contained reference & No dangling pointers such as ``this method'' or ``the figure above''. & $97.17\%$ \\
No internal codes & No enumeration of identifiers meaningful only inside the paper. & $96.95\%$ \\
Single focus & Exactly one answer is required; choosing among listed options counts as one such as ``How does it do on ImageNet, and on COCO?''. & $86.15\%$ \\
Answer shape & The gold matches its declared type: a comparison names a winner, a trend a direction, a regime a condition, and none is a disguised yes/no. & $85.43\%$ \\
\textbf{Abstract-anchored} & A solver shown only the paper's \emph{original} abstract judges that a curious reader would plausibly ask the question, rejecting those reverse-engineered from the evidence. & $\mathbf{26.79\%}$ \\
Phrasing repair & Questions whose wording cannot be made to read naturally are discarded. & $24.79\%$ \\
\textbf{Evidence sufficiency} & A solver given the gold evidence answers correctly. & $\mathbf{22.06\%}$ \\
On-concern & Answering remains informative about the reader's original concern. & $21.90\%$ \\
\textbf{Not memorised} & A tool-free solver fails to produce the gold from parametric knowledge. & $\mathbf{14.91\%}$ \\
Not prior-guessable & Generic ML priors---more data or scale wins, the proposed method wins, ablating a component hurts---do not predict the gold. & $3.35\%$ \\
Not web-findable & A search-and-visit agent fails after exhausting a four-call budget. & $3.35\%$ \\
\textbf{Modality necessity} & On figure- and table-grounded questions, a text-only solver forced through four retrieval rounds over the body text fails. & $\mathbf{2.82\%}$ \\
Modality sufficiency & On text-grounded questions, the same solver succeeds, checking the modality label in the other direction. & $2.33\%$ \\
\bottomrule
\end{tabular}
\caption{Filters applied to drafted questions, in the order they run. \emph{Remaining} is the
share of drafted questions still alive after each filter. The four filters described in
\autoref{sec: benchmark} are in bold.}
\label{tab:filter_funnel}
\end{table}

\subsection{\benchmark Corpus Detail}
We specify the venues of the papers that form \benchmark corpus.
\begin{table}[H]
\centering
\small
\begin{tabular}{lrrrr}
\toprule
 & \multicolumn{2}{c}{Papers} & \multicolumn{2}{c}{Pages} \\
\cmidrule(lr){2-3} \cmidrule(lr){4-5}
Venue & Count & Share & Count & Share \\
\midrule
NeurIPS & 21,448 & 21.7\% & 232,136 & 21.9\% \\
ACL & 12,870 & 13.0\% & 114,720 & 10.8\% \\
CVPR & 10,312 & 10.4\% & 96,869 & 9.1\% \\
ICML & 8,654 & 8.7\% & 92,765 & 8.7\% \\
ICLR & 8,561 & 8.6\% & 86,688 & 8.2\% \\
EMNLP & 7,946 & 8.0\% & 71,194 & 6.7\% \\
ICCV & 5,685 & 5.7\% & 53,913 & 5.1\% \\
AAAI & 4,835 & 4.9\% & 42,178 & 4.0\% \\
NAACL & 1,834 & 1.9\% & 16,317 & 1.5\% \\
ECCV & 1,687 & 1.7\% & 26,515 & 2.5\% \\
Other venues & 8,308 & 8.4\% & 122,281 & 11.5\% \\
\midrule
Preprint only (arXiv) & 6,915 & 7.0\% & 106,577 & 10.0\% \\
\midrule
Total & 99,055 & 100.0\% & 1,062,153 & 100.0\% \\
\bottomrule
\end{tabular}
\caption{Venue composition of the 99,055-document, 1,062,153-page corpus. 93.0\% of the documents appeared at a peer-reviewed venue.}
\label{tab:corpus-venues}
\end{table}

\subsection{\benchmark Annotation}
\begin{table}[H]
\centering
\small
\begin{tabular}{lrr}
\toprule
Outcome & Count & Share \\
\midrule
Accepted as generated & 107 & 10.3\% \\
Kept after editing & 393 & 37.9\% \\
\midrule
Entering \benchmark & 500 & 48.3\% \\
Rejected & 536 & 51.7\% \\
\midrule
Reviewed & 1{,}036 & 100.0\% \\
\bottomrule
\end{tabular}
\caption{Human annotation of the \benchmark. Of $1{,}036$ generated questions reviewed, about one in ten survived untouched and over half were rejected.}
\label{tab:annotation-funnel}
\end{table}

As described in~\autoref{sec: benchmark}, every question that passes all verification gates is reviewed manually before entering \benchmark. We detail the annotation procedure here.

For each question, the annotator sees the question, the answer, the gold page, and the specific grounded block, such as a text span, figure, table, or equation. The full gold paper is also available. The annotator checks three things: (1) the question is well-formed and valid; (2) the answer is concise enough for reliable LLM grading; (3) the gold page is correct and sufficient to answer the question. Annotators may edit a question to fix these issues or reject it.

Only about one in ten questions passes annotation unchanged, and about half are rejected. The main reasons for rejection are that the gold answer contradicts the paper, that no single page suffices to answer the question, or that the answer is too open-ended to grade reliably. Annotation takes about 10 minutes per question.

\subsection{\benchmark Retrieval Difficulty}
\begin{table}[H]
\centering
\small
\begin{tabular}{lrrrrrr}
\toprule
Target & R@1 & R@5 & R@10 & R@30 & R@100 & MRR \\
\midrule
Page & 13.8 & 26.8 & 32.8 & 41.0 & 48.8 & 0.200 \\
\bottomrule
\end{tabular}
\caption{Retrieval difficulty on \benchmark, using Qwen3-VL-Embedding-8B.}
\label{tab:retrieval-difficulty}
\end{table}

\section{\ours Inference Detail}
\subsection{Deployment and Engineering Detail}
\paragraph{Serving.}
GPT-5.4 is accessed through the OpenAI Responses API and Gemini-3.5-Flash through Vertex AI. Qwen3.6-27B, both the base model and our trained variants, is served with SGLang on four H200 GPUs in tensor-parallel mode at a context length of $262{,}144$ tokens. Reasoning is set to \texttt{high} for GPT-5.4 and Gemini-3.5-Flash, and enabled for Qwen3.6-27B. Following the Qwen model card, which warns that greedy decoding in thinking mode induces endless repetition, the open model is sampled at temperature $0.6$, $\text{top-}p=0.95$, $\text{top-}k=20$ with a presence penalty of $1.0$; the proprietary models use provider defaults. All three run under an identical prompt, tool set and search budget.

\paragraph{Halting at the tool call.}
The agent must stop generating the moment it emits a tool call, so that the real retrieval result re-enters the context. For Qwen3.6-27B and Gemini-3.5-Flash we pass \texttt{</call\_tool>} and \texttt{</answer>} as stop sequences. For GPT-5.4 the tools are declared as native functions and the API halts at the \texttt{function\_call} item, with retrieved pages returned as image content in \texttt{function\_call\_output}.

\paragraph{Carrying reasoning across turns.}
 Qwen3.6-27B writes its \texttt{<think>} block in-band, so it is part of the conversation. For the proprietary models we echo the provider's own representation verbatim, the \texttt{thought\_signature} for Gemini-3.5-Flash and the encrypted reasoning item for GPT-5.4, carried in the request body with server-side retention disabled.

\paragraph{Context growth and prefix reuse.}
Every observation stays in context, so the agent can re-read a page it retrieved ten steps earlier, and the append-only prefix is matched by the server's radix cache at every turn, giving a $96\%$ multimodal cache hit rate.

\subsection{Comparison on Retrieval Budget}
\label{sec; retrieval_budget}
In \cref{sec: prelim_exp}, each text search returns $k=5$ chunks of $256$ words, while each \ours search returns $k=3$ pages. To check that this does not give \ours more information, we run OCR on the retrieved pages and compare their text length with that of the retrieved chunks. On average, the text in three retrieved pages is $1.08\times$ that of five retrieved chunks (\autoref{tab:retrieval-budget}), so the two settings receive a comparable text budget per search, and \ours's advantage cannot be attributed to seeing more text from pages.
\begin{table}[H]
\centering
\small
\begin{tabular}{llrr}
\toprule
Setting & Retrieved unit & $k$ & Words per search \\
\midrule
Text        & $256$-word chunk & 5 & 1{,}257 \\
Page-native & page image      & 3 & 1{,}359 \\
\bottomrule
\end{tabular}
\caption{Text returned per search. Word counts are measured over $800$ randomly sampled corpus documents including $22{,}149$ chunks of text, $12{,}296$ pages, with page text obtained by OCR. A chunk averages $251$ words and a page $453$.}
\label{tab:retrieval-budget}
\end{table}

\section{More on \ourstrained}
\label{app:training_setup}
\subsection{\ourstrained Training Detail}

\paragraph{Optimization.}
We fine-tune Qwen3.6-27B with LoRA of rank $32$ and scaling $\alpha=64$, dropout $0.05$, on every linear projection of the language model; the vision encoder stays frozen. Training runs for one epoch at a constant learning rate of $1\mathrm{e}{-4}$ with an effective batch size of $16$, from four H200 GPUs at one sequence per device and four gradient-accumulation steps. We use \texttt{bfloat16}, FlashAttention, Liger kernels and padding-free packing, and set the maximum sequence length to $131{,}072$ tokens.

\paragraph{Applying the weights.}
Counterfactual credits are computed offline against the frozen base policy at two forward passes per step and written into the training file as a per-message weight. Each assistant message is split at the boundary between its reasoning and its action so the two channels can carry different weights, and the trainer multiplies each message's token loss by its weight. The training step is therefore ordinary supervised fine-tuning at no extra cost, and the whole method sits in data preparation. Observation tokens are excluded from the loss in every arm.

\paragraph{Training data.}
Training questions come from the same pipeline as the benchmark, taken before human annotation, over a disjoint pool of papers, and no paper supporting a training question supports an evaluation question. From $1{,}164$ \emph{visual} seed questions whose gold evidence anchored on a table, figure, or equation, we sample six GPT-5.4 rollouts each under \ours, discard those with an incorrect final answer, and keep at most the three shortest trajectories per question, leaving $2{,}500$ trajectories.  The teacher's trajectories are short, $5.1$ searches per question, but heavy-tailed: under uniform token weighting the length seen by the gradient is $E[n^2]/E[n]=9.3$ rather than $5.1$, and trajectories with at least ten searches are $15\%$ of the data but $42\%$ of the steps. Filtering for correctness also means every retained trajectory ends with the answer in its context, which evaluation does not guarantee.

\subsection{\ourstrained Behavior Comparison with GPT-5.4 and \oursbase}
\begin{table}[H]
\centering\small
\setlength{\tabcolsep}{8pt}
\begin{tabular}{@{}lrrrrr@{}}
\toprule
& GPT-5.4 & \multicolumn{2}{c}{\oursbase} & \multicolumn{2}{c}{\ourstrained} \\
\cmidrule(lr){2-2}\cmidrule(lr){3-4}\cmidrule(lr){5-6}
& Score & Score & $\Delta$ & Score & $\Delta$ \\
\midrule
\multicolumn{6}{@{}l}{\textit{Retrieval}} \\
\quad Page Hit                  & 45.7 & 36.3 & $-9.4$  & 52.0 & $+6.3$ \\
\quad Repeated Queries          & 0.0  & 5.3  & $+5.3$  & 0.1  & $+0.1$ \\
\addlinespace
\multicolumn{6}{@{}l}{\textit{Reading}} \\
\quad Acc $\mid$ Page Hit       & 71.2 & 63.9 & $-7.3$  & 67.7 & $-3.5$ \\
\quad\quad from Last Search     & 74.4 & 69.3 & $-5.1$  & 84.2 & $+9.8$ \\
\quad\quad from Earlier Search  & 69.7 & 57.4 & $-12.3$ & 66.9 & $-2.8$ \\
\bottomrule
\end{tabular}

\caption{Decomposing the gap to GPT-5.4 on \benchmark, with $\Delta$ measured against GPT-5.4 in both blocks. \textbf{Page hit} is the share of questions whose gold page is retrieved, and \textbf{repeated queries} the share of search calls that re-issue a query already sent in the same trajectory. \textbf{Acc $\mid$ page hit} is accuracy given the gold page was retrieved, split by whether the page arrived from the last search or had to survive further searches. Training reverses the retrieval gap, and the reading deficit that remains is confined to pages carried from an earlier search.}
\label{tab:gap_decomposition_trained}
\end{table}

\subsection{\ourstrained Training Ablation}
\begin{table}[H]
\centering\small
\setlength{\tabcolsep}{10pt}
\begin{tabular}{@{}lrrrr@{}}
\toprule
Model & Tool/q & PgHit & Acc$\mid$Pg & Acc \\
\midrule
\multicolumn{5}{@{}l}{\textit{Open-weight (Qwen3.6-27B)}} \\
\quad \oursbase                    & 8.8  & 36.3 & 63.9 & 40.1 \\
\quad + vanilla SFT                 & 14.0 & \textbf{53.6} & 67.2 & 49.0 \\
\quad \ourstrained                 & 8.9  & \underline{52.0} & 67.7 & \textbf{52.1} \\
\addlinespace
\multicolumn{5}{@{}l}{\textit{Ablation of the credit}} \\
\quad + answer weighting $\lambda$ & 11.6 & 54.9 & 67.2 & 48.9 \\
\quad + counterfactual credit      & 11.0 & 51.9 & 68.7 & 50.9 \\
\bottomrule
\end{tabular}
\caption{Results on \benchmark. \textbf{Tool/q} is searches per question. \textbf{PgHit} is the percentage of questions for which the gold page is retrieved, and \textbf{Acc$\mid$Pg} is accuracy restricted to those questions. The lower block adds one component of the credit at a time on top of vanilla SFT baseline, and \ourstrained is the full method, which additionally corrects the reasoning and searching channels against each other. Each component removes searches while gradually benefit the final accuracy.}
\label{tab:ablation_training}
\end{table}

\ourstrained adds three ingredients to vanilla SFT: a constant weight $\lambda=1.5$ on the final answer, the per-step counterfactual credit of \autoref{eq:cf} converted to weights by rank, and the cross-channel correction described after \autoref{eq:w}. \autoref{tab:ablation_training} adds them one at a time, with every variation trained on the same $2{,}500$ trajectories under the same recipe, so weighting is the only variable. Each ingredient removes searches, from $14.0$ per question under uniform weighting to $8.9$ under the full method and gradually benefit the final accuracy.

\subsection{Other Variation for \emph{Counterfactual Credit Reweighting}}
\begin{table}[H]
\centering\small
\setlength{\tabcolsep}{10pt}
\begin{tabular}{@{}lrrrr@{}}
\toprule
Model & Tool/q & PgHit & Acc$\mid$Pg & Acc \\
\midrule
\multicolumn{5}{@{}l}{\textit{Reference}} \\
\quad Global rank                  & 11.0 & 51.9 & 68.7 & 50.9 \\
\addlinespace
\multicolumn{5}{@{}l}{\textit{Weight magnitude}} \\
\quad Mean-weight matching         & 10.1 & 50.4 & 63.1 & 46.8 \\
\addlinespace
\multicolumn{5}{@{}l}{\textit{Rank support}} \\
\quad Within-trajectory rank       & 10.0 & 50.3 & 70.9 & 50.1 \\
\quad Within-question rank         & 9.5  & 50.2 & 66.9 & 48.6 \\
\addlinespace
\multicolumn{5}{@{}l}{\textit{Trajectory contribution}} \\
\quad Per-trajectory mass          & 8.9  & 51.6 & 64.0 & 45.6 \\
\bottomrule
\end{tabular}
\caption{Normalising the credit, on \benchmark. Each variant changes one normalisation choice while the credits, the mask and the training recipe stay fixed. Columns are as in \autoref{tab:in_distribution_result_ours}. Rescaling the weights costs reading accuracy, whether the mean weight is matched to the uniform baseline or the mass each trajectory contributes is equalised, while changing the pool a step is ranked against does not help.}
\label{tab:credit_calibration}
\end{table}

\autoref{eq:w} fixes three properties that the counterfactual credit does not itself determine: the magnitude of the surviving weights, the pool of steps a given step is ranked against, and how much mass a single trajectory may contribute. Here we ablate on those three properties in~\autoref{tab:credit_calibration} with the training recipe held fixed.

\paragraph{Weight magnitude.}
Masking every step of non-positive credit zeroes about a quarter of the steps and leaves the mean weight roughly a third below the uniform baseline, so the weighted run effectively takes smaller optimisation steps, and a constant factor on all weights is equivalent to a change of learning rate. Restoring the mean to the baseline settles this: accuracy given the gold page falls from $68.7$ to $63.1$, the largest reading loss in the family, while the number of searches barely moves.

\paragraph{Rank support.}
A counterfactual credit is larger on hard questions, where deleting any step moves the likelihood of the gold answer substantially, so ranking against the whole training set imports question difficulty into a step-level signal. The effect is measurable: the mean surviving weight of a trajectory ranges from $0.65$ at the tenth percentile to $1.70$ at the ninetieth, so a step's weight depends substantially on which trajectory it belongs to. We also try ranking a step only against the other steps of its own trajectory, and only against the steps of the same question's rollouts. Both remove the leak, because a step is then never compared with a step from a different question. Neither helps: within-trajectory ranking stays on par with the reference (50.1 vs.\ 50.9) and within-question ranking is 2.3 points lower, so the bias is real but harmless.

\paragraph{Trajectory contribution.}
Equal mean weights do not equalise what a trajectory contributes: its total weighted mass spans a factor of about six between the tenth and ninetieth percentiles, since a long or lightly masked trajectory supplies far more weighted tokens. Equalising that mass is the most aggressive variant we tried and the weakest result, five points below the reference. This is because a light trajectory gets multiplied by a large ratio.

Numerical calibration is thus not where the credit acts. Rescaling the weights either does nothing or degrades reading, and changing the support of the rank does nothing at all. What remains is which steps the credit masks and how that mask is distributed along a trajectory.

\subsection{\emph{Counterfactual Credit Reweighting} Mechanism Explanation}
\begin{figure}[H]
\centering
\includegraphics[width=\linewidth]{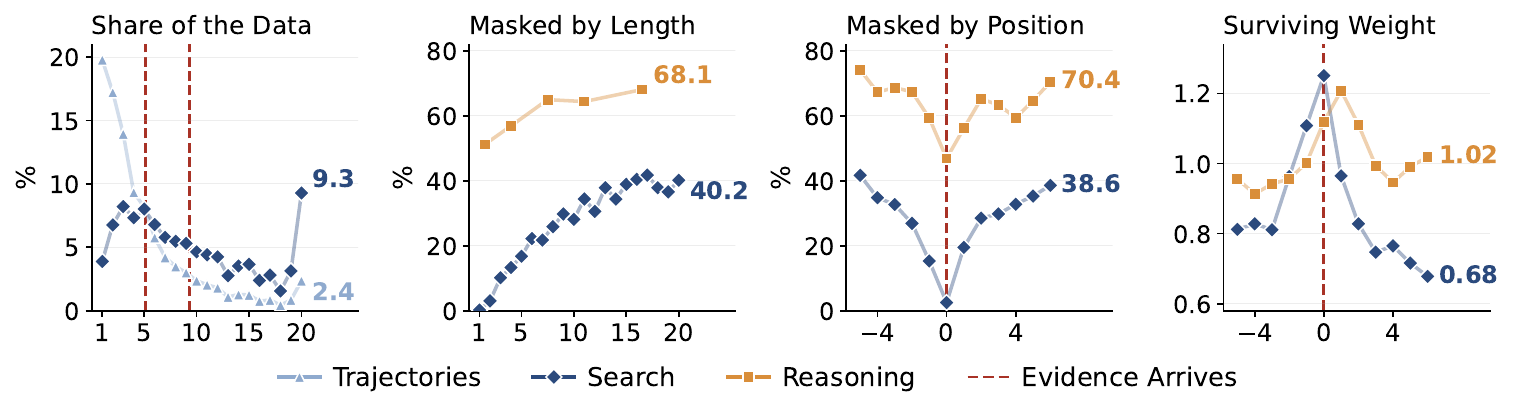}
\caption{What the credit masks in the $2{,}500$ training trajectories. \textbf{Share of the Data}:
trajectories and search steps by trajectory length; The dashed rules mark the average trajectory, which has $5.1$ searches, and the same average weighted by how many steps each trajectory contributes, which is $9.3$. \textbf{Masked by Length}: mask rate against trajectory length. \textbf{Masked by Position} and \textbf{Surviving Weight}: the same two quantities against position relative to the search that first retrieves a page the final answer cites, marked by the dashed rule; that search is masked $2.5\%$ of the time, the minimum of the panel.}
\label{fig:credit_mask}
\end{figure}

\begin{figure}[H]
\centering
\includegraphics[width=0.6\linewidth]{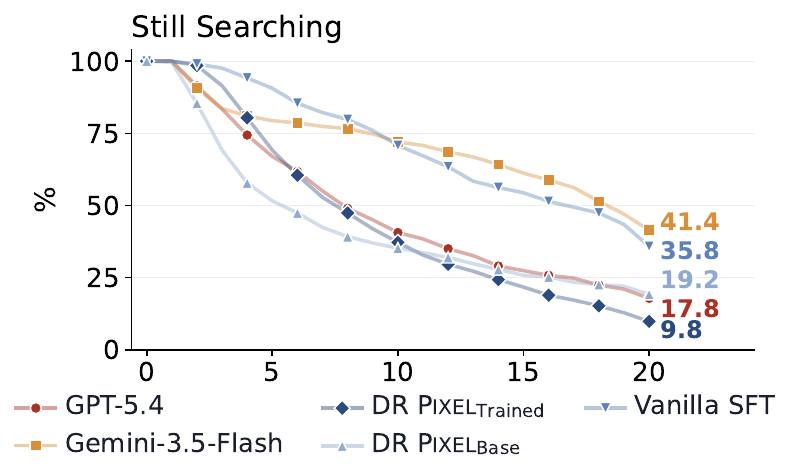}
\caption{Share of questions that issue at least $k$ searches on \benchmark. The harness stops a
trajectory at $20$ calls, so the right-hand endpoint of each curve is that model's exhaustion
rate. \ourstrained is the lowest curve from the twelfth search on and the only model that
ends below $10\%$. Vanilla SFT and Gemini-3.5-Flash share a shape: both stay above $75\%$ for
the first eight searches and then run out of budget rather than deciding to stop.}
\label{fig:search_survival}
\end{figure}

Vanilla SFT and \ourstrained see the same trajectories and differ only in how the loss is weighted, yet one issues $14.0$ searches per question and the other $8.9$ at the same page hit rate. We ask what the two models do differently at inference, and what in the loss produces it.

\paragraph{At inference, the difference is stopping.}
\autoref{fig:search_survival} plots the share of questions still searching after $k$ calls. The curves separate at once and never cross: by the fifth search \ourstrained has finished a third of its questions and vanilla SFT almost none. Per step, averaged over the first twelve searches, \ourstrained stops with probability $10.2\%$, close to the untrained backbone's $9.4\%$ and GPT-5.4's $8.9\%$, while vanilla SFT stops with probability $4.4\%$. At that rate the expected trajectory is $22.8$ searches, longer than the budget allows, so the cap is what ends a third of its trajectories.

\paragraph{The credit is not symmetric across the two channels.}
The credit is much harsher on reasoning than on search. As shown in \autoref{fig:credit_mask}, it masks $62.6\%$ of reasoning steps but only $24.5\%$ of search steps. Only search reacts to trajectory length: almost nothing is masked on trajectories with a single search, $0.2\%$, against $40.2\%$ on trajectories that reach the budget, while reasoning stays between $51\%$ and $68\%$ whatever the length. The small slope reasoning does have is borrowed, since the cross-channel correction of \cref{sec:training_method} masks a reasoning step whenever its query is masked. Length is therefore handled by the search channel: longer trajectories lose a larger share of their search steps, which is how the credit offsets the length bias, and reasoning follows along.

\paragraph{The credit spares the search that finds the evidence and prunes the ones after it.}
In each trajectory we locate the first search that returns a page the final answer cites, and measure the mask rate at each position relative to it. That search is almost never masked, $2.5\%$ of the time, while every other search is masked between $15\%$ and $42\%$. Among the steps that survive it also carries the largest weight, $1.25$, against $0.7$ to $0.8$ for searches a few steps later. Reasoning peaks one step further on, at $1.21$. This position is never supplied to the method, so the pattern is emergent. Because almost every step the credit removes is issued after the evidence is already in context, the pruning suppresses continuation rather than retrieval, which is why the search count falls while the page hit rate does not.

\section{More on Experiment}
\subsection{Supplement Material for Ablation}
\begin{figure}[H]
\centering
\includegraphics[width=0.8\textwidth]{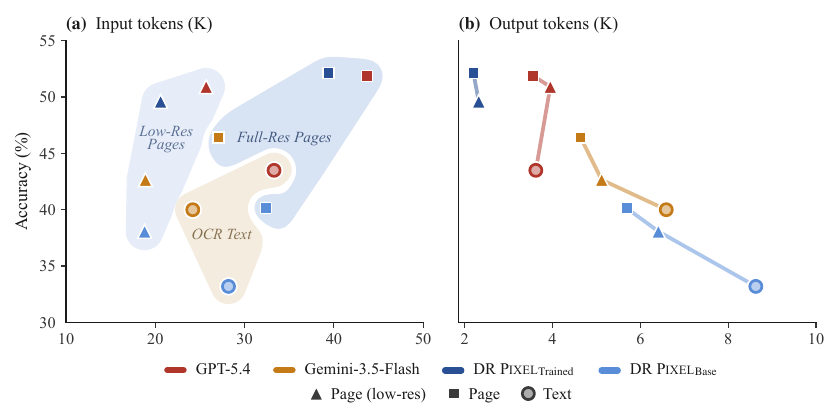}
\caption{Accuracy against (a) input and (b) output tokens as page resolution varies. Lowering resolution cuts input tokens by 30\% to 48\% for a 1 to 4 point drop, while staying 2.7 to 7.4 points more accurate than the OCR-based Text setting.}
\label{fig:accuracy_efficiency_split}
\end{figure}

\section{Prompts}
\begin{figure*}[p]
\begin{promptbox}{Agent system prompt}
\begin{imginput}
{\ttfamily\scriptsize \{retrieved page images, appended to every tool result\}}
\end{imginput}
\vspace{3pt}
\begin{Verbatim}[fontsize=\tiny, baselinestretch=0.92]
You are a research assistant who answers questions about scientific papers
through iterative reasoning and research.

## Process
- Show your reasoning inside <think></think> tags.
- Call a tool with exactly:  <call_tool name="TOOL_NAME">your query</call_tool>
- When you have enough evidence, give the final answer with exactly:
  <answer>your answer</answer>
- You may alternate between thinking and tool calls several times.

## STOP after the tag  (this is the most important rule)
Each turn you emit AT MOST ONE <call_tool> or ONE <answer>, and then you STOP WRITING.
Do NOT write a <tool_output> block yourself -- WE run the tool and send you the real
results. Anything you write after the closing </call_tool> is discarded, and any tool
output you invent is a hallucination: the documents, page numbers and numbers in it
are not real.

Your turn looks like exactly this and nothing more:

    <think>What I know so far, and what I still need.</think>
    <call_tool name="TOOL_NAME">my query</call_tool>

Your <think> stays visible to you for the rest of the session, so record in it any
concrete value you read off an image -- that is your working memory.

## Tools (use ONLY these)
{tools}

## Tool Output
After each tool call the results come back to you in a <tool_output> block:
{obs}
Results from EVERY tool call stay visible to you for the rest of the session -- you
can re-read an image you retrieved earlier.

## Before you answer -- check yourself
You may only answer when you can point to the SPECIFIC page/passage id you read the
answer off (e.g. "P2_1"). Ask yourself:
  - Is the exact value/claim the question asks for actually VISIBLE in one of the
    results I received?
  - If NO -> do NOT answer. Search again with a different query.
  - "The retrieved page is about the right topic" is NOT enough. "I remember this
    paper" is NOT enough.
Answering from memory when the evidence is not in front of you is the single worst
failure mode here, because the answer will look confident and be about the wrong paper.

## CRITICAL ANSWERING RULES
- Answer ONLY from evidence you actually retrieved this session. Do NOT answer from
  prior knowledge.
- When the answer involves a specific number, score, magnitude, threshold, or a named
  model/method/dataset, you MUST retrieve it and state the SPECIFIC value(s). A vague
  or directional answer ("generally better", "tends to improve", "competitive",
  "around X") when a precise value exists is treated as WRONG.
- NEVER guess or fabricate numbers, model names, method names, arxiv ids or document
  ids. If after searching you still cannot find the specific value the question asks
  for, say so explicitly -- do NOT invent one.
- You start this session having retrieved NOTHING. The ONLY evidence that exists is
  what comes back inside a real <tool_output> block that WE send you. If you have not
  yet received a <tool_output>, you have no evidence and you MUST NOT answer -- search
  first.
- Recalling the paper from memory is not evidence. If your answer would be the same
  without looking at the retrieved pages, you have not done the task.
- For yes/no, comparison, trend, or regime questions: state the verdict/direction AND
  the supporting numbers behind it.

Illustrative GOOD vs BAD answers (fabricated examples, NOT the real question):
- Q "Does pruning hurt accuracy on the small model?"
  BAD : "Yes, pruning generally lowers accuracy somewhat."        (vague -> WRONG)
  GOOD: "Yes -- accuracy drops from 79.3% to 74.1% at 50% sparsity (Table 3), a
         5.2-point fall."
- Q "Which retriever wins on the multilingual benchmark, and by how much?"
  BAD : "A single general retriever is competitive across languages."
                                                            (names nothing -> WRONG)
  GOOD: "Retriever-Z leads, +1.8 nDCG@10 over the per-language baseline (Table 4)."

Rules: you may call tools at most {budget} times in total; after that you MUST answer.
Issue focused queries. Read the returned passages/images carefully.
\end{Verbatim}
\end{promptbox}
\caption{The agent system prompt, identical for every model we evaluate.
\texttt{\{tools\}}, \texttt{\{obs\}} and \texttt{\{budget\}} are filled at run time,
with $\texttt{budget}=20$. Models whose tool call is a native API item rather than a tag
receive the same text with only the \textsc{Process} block replaced.}
\label{fig:prompt_agent}
\end{figure*}

\begin{figure*}[p]
\begin{promptbox}{Answer grading prompt}
\begin{Verbatim}[fontsize=\scriptsize]
You STRICTLY grade an answer to a question about a research paper's figure/table/finding.
The GOLD states a SPECIFIC finding one can only know by reading the figure/table -- often
exact numbers, named models/methods/datasets, or a precise quantitative relationship.

GRADING RULE:
1. If the GOLD contains specific numbers, named entities (models/methods/datasets), or a
   precise magnitude, mark correct=true ONLY IF the PREDICTED answer states that SAME
   specific finding: the same numbers (within rounding) AND/OR the same named entities
   AND/OR the same precise relationship. Otherwise correct=FALSE. In particular:
   - Generic / directional answer that gets the trend right but OMITS the gold's specific
     numbers => WRONG.
   - Answer citing DIFFERENT numbers or DIFFERENT models/methods than the gold => WRONG
     (it guessed from prior knowledge, did not read the source).
   - Vague hedging ("levels off", "tends to", "generally", "competitive") in place of the
     gold's specific finding => WRONG.
2. If the GOLD has NO specific numbers/entities (a pure yes/no verdict or a purely
   qualitative direction), then a correctly matching verdict/direction is enough.
Paraphrase of the SAME specific finding is fine; extra correct detail is fine; a
contradicted or reversed direction is always WRONG.

answer_type = {atype} (verdict / comparison / trend / regime).

CALIBRATION EXAMPLES:
[CORRECT] GOLD: "No, it stays stable..." (no numbers) | PRED: "No, ... remains bounded,
  avoids collapse" -> verdict matches, gold has no specifics.
[CORRECT] GOLD: "operational factors; limiters = latent size and number of modules" |
  PRED: "operational overhead; size and number of processed tensors" -> states the same
  specific finding.
[WRONG] GOLD: "Jina-ColBERT-v2 within ~0.4 nDCG@10 of mDPR-FT on MIRACL" | PRED: "a single
  model such as CORA or mAggretriever is competitive" -> different models, none of the
  gold's specifics.
[WRONG] GOLD: "EMOVA 72B trails on RealWorldQA 71.0 vs 75.4" | PRED: "InternVL3-78B,
  Qwen-VL2.5-72B trail GPT-4o" -> different models, missing 71.0/75.4.
[WRONG] GOLD: "edge vanishes: 66.6 vs 62.7 (Qwen2.5) but 70.3 vs 70.9 (Qwen3)" | PRED:
  "the advantage diminishes" -> correct direction but NONE of the gold's specific numbers.
[WRONG] GOLD: "rises 2->4, peaks ~4, dips at 5" | PRED: "levels off / plateaus" -> vague,
  misses the specific peak/dip.

QUESTION: {q}
GOLD ANSWER: {gold}
PREDICTED ANSWER: {pred}
First state the gold's KEY specific finding, then decide.
Output ONLY: {"gold_key_finding": "<the specific finding the gold makes>",
              "correct": true or false,
              "reason": "<one sentence: does PRED state it?>"}
\end{Verbatim}
\end{promptbox}
\caption{The grading prompt. A prediction is scored by majority of three independent votes
from GPT-5.4-mini; an empty prediction is marked incorrect without being sent to the judge.
Retrieval metrics do not pass through this prompt: page hit and document hit are decided by
exact identifier match against the gold annotation.}
\label{fig:prompt_judge}
\end{figure*}

\begin{figure*}[p]
\begin{promptbox}{Atomic fact extraction prompt (figure modality)}
\begin{imginput}
{\ttfamily\scriptsize \{cropped figure image\}}
\end{imginput}
\vspace{3pt}
\begin{Verbatim}[fontsize=\scriptsize]
You are extracting fine-grained factual content from a scientific paper FIGURE (the image).
You are given the figure image, its caption, and the surrounding body text.

Return STRICT JSON:
{
 "overview": "<=2 sentences on what the figure shows",
 "atomic_facts": [
   {"fact":"<one concrete, checkable fact READ FROM THE IMAGE>",
    "fact_type":"trend|value|comparison|ranking|structure|annotation",
    "stated_in_caption": true|false,
    "stated_in_nearby_text": true|false,
    "salience": 1-5}
 ]
}
Rules:
- Facts must be things visible IN THE IMAGE PIXELS (values, trends, orderings, crossover
  points, error bars, sub-figure differences, color/legend mappings).
- For each fact, judge honestly whether the SAME information is already explicitly stated
  in the caption / nearby text provided.
- Prefer facts that are NOT stated in caption/text. 4-10 facts.

CAPTION: {caption}

NEARBY TEXT: {nearby}
\end{Verbatim}
\end{promptbox}
\caption{Atomic fact extraction, shown for the figure modality. Tables receive the same
prompt with the OCR'd cell text in place of the image; equations receive the \LaTeX{} source
with no image; and body text is processed in chunks without caption or salience fields. The
\texttt{stated\_in\_caption} and \texttt{stated\_in\_nearby\_text} flags are what later let us
keep only the facts a reader could not obtain from the surrounding prose.}
\label{fig:prompt_facts}
\end{figure*}

\begin{figure*}[p]
\begin{promptbox}{Question generation prompt}
\begin{Verbatim}[fontsize=\scriptsize]
You read this (augmented) abstract. As a working researcher, pick ONE thing you'd genuinely
want to check whose precise answer needs a figure/table/equation (a gap / head-to-head /
where-it-breaks / dependency / feasibility / exception) -- NOT a vague survey question, NOT
a trivial lookup.
BE IMAGINATIVE -- go for a NON-OBVIOUS, slightly surprising angle a curious reader of the
ABSTRACT would wonder about: an unexpected regime where the claim might break, a
counterintuitive comparison, a "what's the catch", a hidden tradeoff. Avoid the rote "does
the proposed method beat the baseline?". The thing you ask about must be something the
ABSTRACT itself raises (a claim/effect/tradeoff) -- NOT an internal experimental knob the
abstract never mentions. Express it as an expectation ("I'd expect X because Y"), pick the
answer_type, give 2-3 grep keywords appearing in the paper's figure/table facts
(dataset/method/metric words).{modhint}

AUGMENTED ABSTRACT:
{aug}

Return JSON {"folk_belief":"...","intent":"...",
             "answer_type":"comparison|trend|regime|verdict",
             "keywords":["..","..",".."]}
\end{Verbatim}
\end{promptbox}
\caption{Question generation. The model sees only the abstract, enriched beforehand with the
atomic facts of Figure~\ref{fig:prompt_facts}, and never the body of the paper, so the
question is driven by what a reader would wonder rather than by evidence already in view.
Two further prompts, omitted here, turn the resulting expectation into a precise question
with a short gold answer and then rewrite it in a researcher's voice.}
\label{fig:prompt_qgen}
\end{figure*}

\end{document}